\pdfoutput=1

\documentclass[11pt]{article}
\usepackage{CJKutf8}
\usepackage{colortbl}
\usepackage[table]{xcolor}
\usepackage{amsmath}
\usepackage{graphicx}
\usepackage{booktabs} 
\usepackage{array} 
\usepackage{acl}
\usepackage{times}
\usepackage{latexsym}
\usepackage{graphicx}
\usepackage{eso-pic}
\usepackage{float}
\usepackage{multirow}
\usepackage[T1]{fontenc}

\usepackage[utf8]{inputenc}
\usepackage{pifont}
\usepackage{microtype}

\usepackage{inconsolata}

\usepackage[textsize=scriptsize]{todonotes}

\title{The Impact of Visual Information in Chinese Characters:\\ Evaluating Large Models' Ability to Recognize and Utilize Radicals}

\author{
  Xiaofeng Wu$^1$  \quad Karl Stratos$^2$ \quad Wei Xu$^1$  \\
  $^1$Georgia Institute of Technology  \quad $^2$Rutgers University\\
  \texttt{xwu414@gatech.edu}, \texttt{karl.stratos@rutgers.edu}, \texttt{wei.xu@cc.gatech.edu} 
}

\begin{document}

\definecolor{lightblue}{RGB}{215, 233, 248}
\definecolor{darkred}{RGB}{179, 0, 0}
\definecolor{lightorange}{rgb}{1.0, 0.75, 0.5}
\definecolor{darkgreen}{rgb}{0.23, 0.49, 0.14}
\definecolor{darkyellow}{rgb}{1.0, 0.75, 0.0}
\definecolor{darkpink}{rgb}{0.90, 0.62, 0.87}

\definecolor{lightgreen}{rgb}{0.7608, 0.8980, 0.8039}
\definecolor{mediumgreen}{rgb}{0.5765, 0.8235, 0.6431}
\definecolor{stronggreen}{rgb}{0.4941, 0.7882, 0.5725}
\definecolor{verydarkgreen}{rgb}{0.3882, 0.7451, 0.4824}
\definecolor{extremedarkgreen}{rgb}{0.1176, 0.6901, 0.2745}
\maketitle
\begin{abstract}
The glyphic writing system of Chinese incorporates information-rich visual features in each character, such as radicals that provide hints about meaning or pronunciation.
However, there has been no investigation into whether contemporary Large Language Models (LLMs) and Vision-Language Models (VLMs) can harness these sub-character features in Chinese through prompting. In this study, we establish a benchmark\footnote{Our benchmark can be accessed through \href{https://github.gatech.edu/xwu414/Chinese-SubCharacter-Dataset.git}{GitHub}.} to evaluate LLMs' and VLMs' understanding of visual elements in Chinese characters, including radicals, composition structures, strokes, and stroke counts. Our results reveal that models surprisingly exhibit some, but still limited, knowledge of the visual information, regardless of whether images of characters are provided.
To incite models' ability to use radicals, we further experiment with incorporating radicals into the prompts for Chinese language processing (CLP) tasks. We observe consistent improvement in Part-Of-Speech tagging when providing additional information about radicals, suggesting the potential to enhance CLP by integrating sub-character information.


\end{abstract}

\section{Introduction}

\begin{CJK*}{UTF8}{gbsn}

\begin{figure}[!t]
    \centering
    \includegraphics[width=7.4cm]{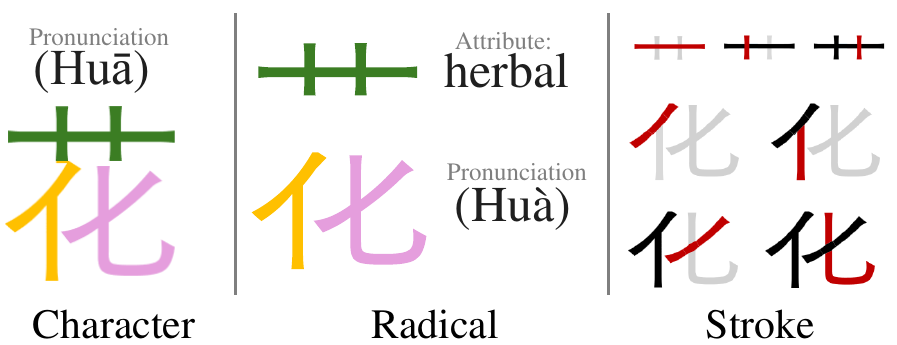}
    \caption{Chinese character ``花'' displayed at the character, radical, and stroke levels from left to right. Different radicals are shown in \textcolor{darkgreen}{\textbf{green}}, \textcolor{darkyellow}{\textbf{yellow}}, and \textcolor{darkpink}{\textbf{pink}} colors, while the writing order of the strokes is indicated by \textcolor{darkred}{\textbf{red}} (current), \textcolor{gray}{\textbf{gray}} (upcoming), and \textcolor{black}{\textbf{black}} (completed).}
    \label{fig:1}
\end{figure}

Visual information embedded in Chinese characters is important, as most Chinese characters convey a meaning equivalent to an entire word in English with a complex glyphic structure. Multiple writing strokes form the \textit{radicals},\footnote{A comprehensive definition of Chinese radicals can be found on Wikipedia: \url{https://en.wikipedia.org/wiki/Chinese_character_radicals}. For simplicity, this paper refers to any large components within a character as radicals.} which often carry information about semantic meaning and pronunciation; the radicals are then visually combined to form Chinese characters. 
When encountering unfamiliar characters, Chinese speakers rely on semantic and phonetic hints from radicals, much like how English speakers use sub-words such as prefixes or suffixes, to approximate the meaning and pronunciation of unfamiliar words. For example, the Chinese character ``花'' (meaning ``flower''; pronounces as ``huā'') in Figure \ref{fig:1} has ``艹'' (meaning ``herbal'') on the top, contributing to its semantic meaning, and ``化'' (pronounces as ``huà'') on the bottom, indicating its pronunciation. By utilizing the radical information, one can infer that ``花'' is related to herbs and has a pronunciation similar to ``huà'' without prior knowledge of the character. 

 
Although radicals contain rich information, they have received little attention in digital text processing. Contemporary typeface treat Chinese characters, radicals, and strokes as indivisible units, disregarding their compositional relationships. Consequently, most language models follow this approach, under-utilizing the rich visual and semantic information embedded in Chinese characters. While limited prior works \cite{sun2021chinesebert, si2021subcharacter, stratos2017subcharacter} have attempted to address this issue by incorporating visual embeddings, such as strokes or font images, into smaller-scale models (i.e., BERT or Neural MT), there remains a lack of research investigating whether these visual features can be recognized and utilized by models in light of the significant advancements in LLMs and VLMs, especially in inference methods (e.g., prompting).

\NewDocumentCommand\emojrad{}{\includegraphics[scale=0.06]{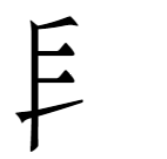}}

\begin{figure}[t]
    \centering
    \includegraphics[width=1\linewidth]{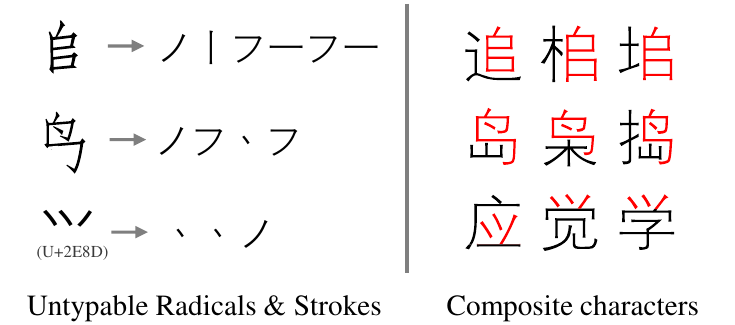}
    \caption{Example images of untypable radicals alongside example characters and their corresponding Unicode values, if available.}
    \label{fig:untypable}
\end{figure}

To determine whether pre-trained LLMs recognize or can acquire the visual knowledge embedded in Chinese characters, we establish a benchmark  by collecting over 14,000 Chinese characters from the Chinese, Japanese, and Korean (CJK) Unified Ideographs,\footnote{The CJK Unified Ideographs refers to a set of Chinese characters used across Chinese, Japanese, and Korean languages to standardize and unify the use of characters.} considering four visual elements: radicals, composition structures, strokes, and stroke counts. As shown in Figure \ref{fig:dataset}, the \textit{composition structure} refers to the visual arrangement of a character’s radicals (e.g., top-to-bottom or left-to-right). \textit{Stroke composition} provide an essential way to represent not typable radicals; 
As shown in Figure \ref{fig:untypable}, some radicals cannot be typed using standard input methods but can still be accurately depicted through stroke compositions (see more details in Appendix \ref{subsec:discussencoding}).
Lastly, \textit{Stroke count} offers a measure for Chinese characters’ visual complexity,  similar to word length in English.   


We conduct experiments on four tasks: structure recognition, radical recognition, stroke count identification, and stroke identification. We evaluated a series of LLMs and VLMs (e.g.,  GPT-4, Gemini-1.5, Ernie-4, Aya-command, QWen-7B, etc.), and found that all models possess some of visual knowledge of Chinese characters, even without image inputs; however, it is only to a limited extent. In particular, the models tend to perform well in recognizing the first radical of a Chinese character, such as ``艹'' (herbal) in ``花''(flower), but often fail with subsequent ones. We also demonstrate that the pixel-based encoder PIXEL \cite{razzhigaev-etal-2022-pixel} has the ability to capture structural information effectively after fine-tuning. As a language model pre-trained only on an English corpus, PIXEL achieved an F1 score of 84.57, significantly higher than the second-best score of 54.30 achieved by Ernie-4\footnote{Released by Chinese company Baidu that rivals GPT-4.}  and 23.29 by GPT-4 when provided with images of characters, indicating its potential for CLP as it naturally captures visual information.

\end{CJK*}
We further investigate whether models can utilize radicals to improve performance on understanding tasks (e.g., POS tagging and NER) by prompting them to use radicals when encountering unfamiliar words. Our experiments show that radical information yields promising results in downstream tasks, particularly in POS tagging. We observe consistent improvement across models and datasets when the information about radicals are provided.  Notably, Ernie-Lite-8K's POS tagging F1 score on GSD\cite{gsdsimp2023} decreases by 2.1 points when recognizing radicals on its own, but increases by 5.7 points when provided with correct radicals. 
For Name Entity Recognition (NER), We also observe an improvement on three out of six models. Analyzing the cases where incorporating radical degrades the model performance, we see that incorrect answers often occur when the model fails to identify unfamiliar words and bypasses the radical information process, indicating the decrease is likely due to long prompts. When evaluating only sentences where the model detects unfamiliar words, performance on NER generally improves. Our work demonstrates that models possess ability to recognize and utilize radical information only to a certain limit, suggesting that deeper integration, such as additional training on radicals or improvement in Chinese digital system to incorporate radical, could unlock further potential.

\section{Related Work}

\paragraph{Chinese Character Decomposition in Computer Vision.}
The task of decomposing Chinese characters into constituent components has majorly been studied in the field of computer vision. Research within this domain, such as the studies by \citet{ma2dgmmhmm}, \citet{xia1994knowledge}, and \citet{liu2021representations}, has explored analogous challenges. The work by \citet{zhang2018radical} employs a methodical approach by categorizing characters into structured types and further decomposing sub-components according to their spatial arrangements—akin to the layered structural analysis which we adopt in this paper.

\paragraph{Chinese Decomposition Datasets.}
\begin{CJK*}{UTF8}{gbsn}
A comprehensive dataset \cite{cjkvi_ids} that offers decompositions for the CJK Unified Ideographs. Although this collection overlaps with our dataset, it does not cite any authoritative sources for its data. This omission leads to ambiguity due to multiple decomposition sequences for individual characters. Our approach utilizes sources from authoritative Chinese dictionaries, such as the Kangxi Dictionary (康熙字典) and the Xinhua Dictionary (新华字典), ensuring a validated framework for visual information. Additionally, our dataset contains standard stroke orders for all 14,648 characters, which the aforementioned dataset lacks. 

\paragraph{Glyphic Embedding Strategies in LMs.}

Few prior works have utilized the idea of adding additional input embedding with Chinese visual features. For instance, \cite{shi2015radicalembedding} attempted to add radical embedding in the pre-transformer era. \cite{sun2021chinesebert} introduced font images into embedding, and \cite{si2021subcharacter} experimented with stroke  among other glyph-based embeddings such as Cangjie\footnote{glyph-based Chinese character input method.} (仓颉). Another interesting approach is PIXEL \cite{razzhigaev-etal-2022-pixel}, which uses a pixel-based encoder to transform input into images. Our experiment results in \S \ref{sec:intrinsic} highlights the potential of pixel-based language models in CLP.

\end{CJK*}

\section{Chinese Character Dataset (CCD)}
\begin{CJK*}{UTF8}{gbsn}
To evaluate contemporary LLMs and VLMs' proficiency with visual information in Chinese characters, we compile a dataset using characters from CJK Unified Ideographs with visual features collected from the digitized Kangxi Dictionary and Xinhua Dictionary. Our dataset includes 14,648 Chinese characters and details their corresponding radicals, strokes, and stroke count. A subset of 4,651 Simplified Chinese characters also contains structural composition information. The detailed statistics are provided in Table \ref{tab:dataset_stats} with three tiers of character frequency based on the Table of General Standard Chinese Characters (通用规范汉字表).\footnote{\href{https://zh.wikisource.org/wiki/\%E9\%80\%9A\%E7\%94\%A8\%E8\%A7\%84\%E8\%8C\%83\%E6\%B1\%89\%E5\%AD\%97\%E8\%A1\%A8}{https://zh.wikisource.org/wiki/通用规范汉字表}}

\begin{figure*}
    \centering
    \includegraphics[width=1\linewidth]{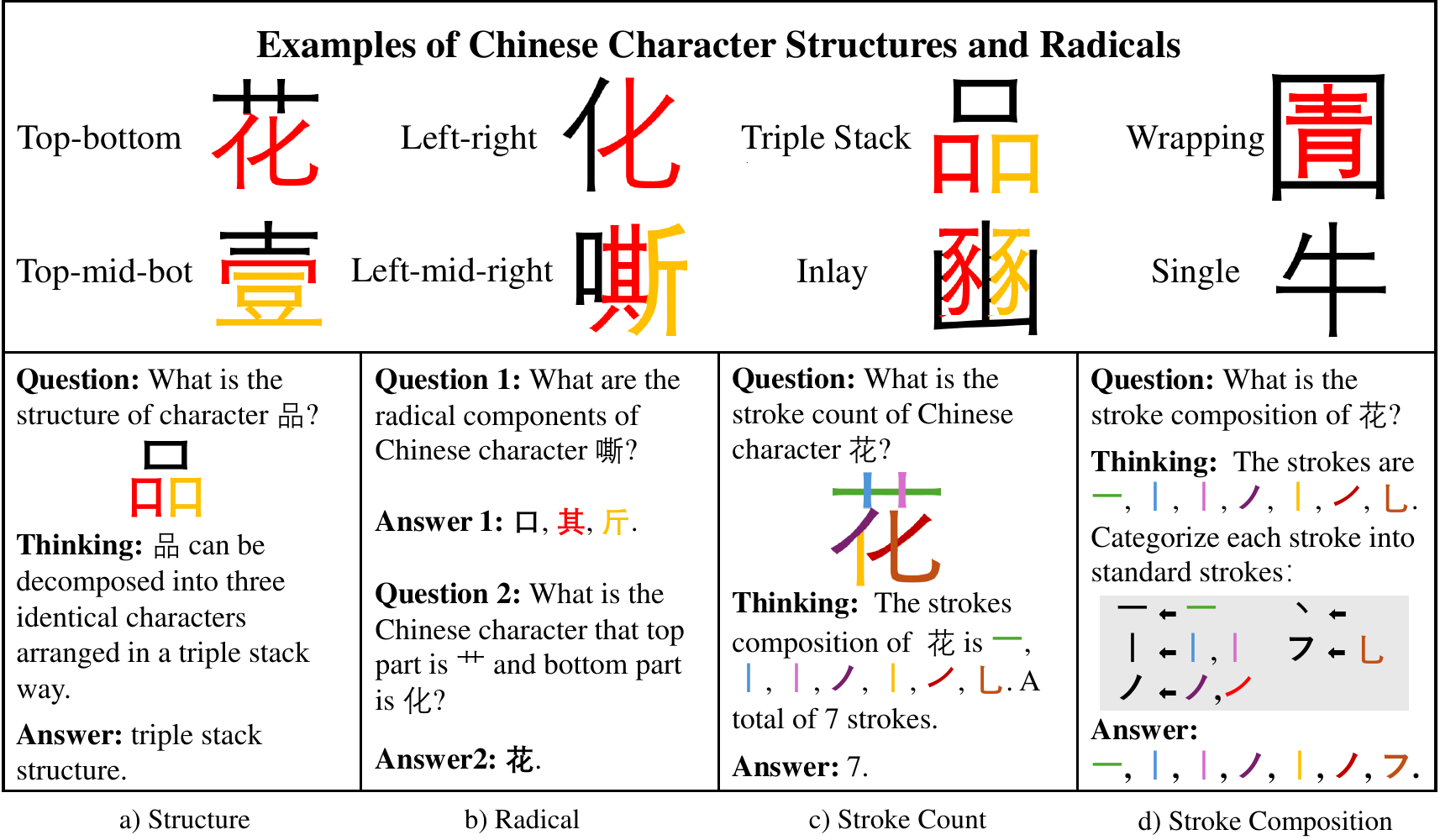}
    \caption{Examples of composition structures with radicals in order of \textbf{black}, \textcolor{red}{\textbf{red}}, \textcolor{darkyellow}{\textbf{yellow}} and four different tasks.}
    \label{fig:dataset}
\end{figure*}

\paragraph{Structure of Chinese Characters.}
According to the digitized Kangxi Dictionary,\footnote{ \url{https://www.kangxizidian.com/}} we categorize 4651 simplified Chinese characters into eight major structural arrangements, with examples of each structure illustrated in Figure \ref{fig:dataset}: top-bottom, left-right, top-mid-bottom, left-mid-right, wrapping, inlay, triple-stack, as well as single structure, which refers to characters that cannot be further segmented. The structure of Chinese characters can be rather complex. For example, the character ``花'', shown in Figure \ref{fig:1}, has a top-bottom structure, consisting of ``艹'' and ``化''. Furthermore, ``化'' exhibits a left-right structure which can be further decomposed into ``亻'' and ``七.'' For the sake of clarity, we categorize all characters based on their top-level structure (i.e., top-bottom for ``花'').

\paragraph{Radicals of Chinese Characters.}
Based on the aformenttioned structural composition, we collect radicals by decomposing each Chinese character into meaningful components that have corresponding Unicode representations. The radicals are then ordered according to specific rules: from top to bottom, left to right, outside to inside, and main parts before inlay parts, as illustrated in Figure \ref{fig:dataset}. We avoid decomposition for simple Chinese characters that would reduce to meaningless strokes. For example, while the character “八” could be segmented as a left-right structure, we classify it as a single structure with only one radical to prevent it from being reduced to meaningless strokes. 

The initial collection of radicals was performed using APISpace's Chinese character segmentation API,\footnote{API documentation can be accessed at \url{https://www.apispace.com/eolink/api/dfsdfsfsf/apiDocument}} which analyzes characters' stroke compositions and extracts radicals based on the optimal sub-sequences of strokes. After the automated annotation, we conducted a thorough two-round manual review to ensure accuracy. More than 1,000 characters required manual adjustments due to missing or incorrect radicals; in addition, another 500 adjustments were made to prevent unnecessary reduction of characters into strokes by one of the native Chinese-speaking authors. Four native Chinese speakers further reviewed the entire dataset and collectively corrected for about 2\% annotations. Further details are provided in the Appendix \ref{sec:annotation}.

\begin{table}[t!]
\centering
\fontsize{10}{10}\selectfont
\begin{tabular}{lr}
\toprule

\textit{\textbf{Character Level}}  & \textbf{Statistic} \\
\midrule
\# of total Chinese characters & 14,648 \\
 - Commonly used (tier 1): & 3,500 (24.1\%) \\
 - Less commonly used (tier 2): & 3,000 (20.6\%) \\
 - Terminology used (tier 3): & 1,605 (11.0\%) \\
 - Hardly ever used (others): & 5,543 (37.8\%) \\
\midrule
 - w/ structure information: & 4,651 (31.8\%) \\
\midrule
\textit{\textbf{Radical Level}} \\
\midrule
\# of unique radical & 2,132\\
\# of single-appearance radical & 692\\
\midrule
\textit{\textbf{Stroke Level}} \\
\midrule
\# of unique stroke composition & 13,740 \\

\# of strokes per character ($\mu$) & 11.51 \\
\# of strokes per character ($\sigma$) & 3.92 \\
\# of strokes per character (min)& 1 \\
\# of strokes per character (max)& 39 \\
\bottomrule
\end{tabular}
\caption{Key statistics of our Chinese character dataset. }
\label{tab:dataset_stats}
\end{table}

\paragraph{Stroke Composition of Chinese Characters.}
Stroke composition refers to the sequence of writing order of a Chinese character's strokes. Chinese dictionaries categorize all Chinese strokes into five basic stroke types: ``一 '', ``丨 '', ``ノ '', ``丶'', and ``フ', which our dataset adopts. We first utilized the Xinhua Dictionary API to automatically annotate. For characters not found in the API, we attempted to concatenate the stroke composition of their radicals in order. We then manually reviewed all stroke compositions to ensure accuracy.
\paragraph{Stroke Count of Chinese Characters.}
 The number of strokes required to write a Chinese character is also present in our dataset, which offers a measure of word complexity. Unlike alphabetic writing system, where word length can hint at complexity, Chinese characters occupy a uniform length of one, making stroke count a valuable indicator of intricacy. The statistics for strokes are provided in Table \ref{tab:dataset_stats} with illustrations in Figure \ref{fig:dataset}.
\end{CJK*}

\NewDocumentCommand\emojwen{}{\includegraphics[scale=0.06]{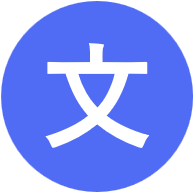}}

\begin{table*}[ht]
\centering
\setlength{\tabcolsep}{3.5pt}
\fontsize{12}{1}
\resizebox{\textwidth}{!}{
\begin{tabular}{lcc|c@{\hskip8pt}c@{\hskip8pt}c@{\hskip8pt}c@{\hskip8pt}c|c|c@{\hskip8pt}c|c@{\hskip8pt}c@{\hskip8pt}c@{\hskip8pt}c@{\hskip8pt}c}
\toprule
\multirow{3}{*}{\textbf{Model}} & \multicolumn{2}{c|}{\textbf{Structure}}& \multicolumn{6}{c|}{\textbf{Radicals}} & \multicolumn{2}{c|}{\textbf{Stroke Ct.}} &
\multicolumn{5}{c}{\textbf{Stroke Composition}}
\\ 

\cmidrule(lr){2-3} \cmidrule(lr){4-9} \cmidrule(lr){10-11} \cmidrule(lr){12-16}
 & \textbf{F1} & \textbf{\(\mathbf{H}\)} & \textbf{1st} & \textbf{2nd} & \textbf{3rd} & \textbf{F1} & \textbf{\(\mathbf{H}\)} & \textbf{Acc} & \textbf{MSE} & \textbf{MAE} & \textbf{1st} & \textbf{2nd} & \textbf{3rd} & \textbf{F1} &\textbf{\(\mathbf{H}\)} \\
 \cmidrule(lr){4-4} \cmidrule(lr){5-5} \cmidrule(lr){6-6} \cmidrule(lr){12-12} \cmidrule(lr){13-13} \cmidrule(lr){14-14}
& \(\boldsymbol{\uparrow}\) & \(\boldsymbol{\downarrow}\) & \textbf{Acc} & \textbf{Acc} & \textbf{Acc} & \(\boldsymbol{\uparrow}\) & \(\boldsymbol{\downarrow}\) & \(\boldsymbol{\uparrow}\) & \(\boldsymbol{\downarrow}\) & \(\boldsymbol{\downarrow}\) & \textbf{Acc} & \textbf{Acc} & \textbf{Acc} & \(\boldsymbol{\uparrow}\) & \(\boldsymbol{\downarrow}\)
 \\
\midrule
\multicolumn{15}{c}{\textit{Vision Language Models (VLMs)}} \\
\midrule
 \(\emojwen{\sc}{}\)Ernie-4& \cellcolor{lightblue}54.30 & - & \cellcolor{lightblue}41.03 & \cellcolor{mediumgreen}34.21 & \cellcolor{mediumgreen}12.50 & \cellcolor{lightblue}41.67 & - & \cellcolor{mediumgreen}71.79 & 12.54 & 1.78 & \cellcolor{mediumgreen}53.85 & \cellcolor{lightblue}35.90 & \cellcolor{mediumgreen}47.37 & \cellcolor{mediumgreen}30.90 & -\\
 \(\emojwen{\sc}{}\)Kimi-v1  & 45.60 & - & 36.73 & 19.15 & 0.00 & 32.93 & - & 42.86 & 15.32 & 2.68 & 30.61 & 26.53 & 16.67 & 20.70 & -\\
Claude-3 & 23.70 & 0.54 & 8.80 & 0.61 & 0.00 & 2.44 & 1.09 & 57.30 & \cellcolor{mediumgreen}5.93 & \cellcolor{mediumgreen}1.22 & 15.40 & 19.60 & 26.80 & 19.62 & 1.22\\
Gemini-1.5 & 27.15 & 0.36 & 3.00 & 0.41 & 0.00 & 1.53 & 1.12 & 27.08 & 8.83 & 2.28 & 29.60 & 16.80 & 22.00 & 22.04 & 1.00\\
GPT-4 & 23.28 & 0.46 & 10.20 & 0.41 & 0.00 & 9.22 & 0.95 & 24.18 & 7.96 & 1.64 & 24.00 & 19.60 & 23.80 & 21.96 & 1.34\\
GPT-4o & 26.66 & - & 6.00 & 0.40 & 0.00 & 6.62 & 0.74 & 67.39 & 10.35 & 1.72 & 46.40 & 31.40 & 34.80 & 30.85 & 0.70\\

\midrule
\multicolumn{15}{c}{\textit{Close-Sourced Models (LLMs)}} \\
\midrule
 \(\emojwen{\sc}{}\)Ernie-Lite-8K& 7.19 & 0.76 & 18.92 & 3.52 & 0.13 & 11.99 & 1.89 & 3.72 & 44.53 & 5.34 & 29.30 & 23.28 & 20.78 & 23.34 & 1.11\\
 \(\emojwen{\sc}{}\)Kimi-v1& 24.51 & 0.83 & 7.24 & 0.33 & 0.00 & 1.10 & 0.72 & 50.16 & 19.05 & 3.12 & 33.12 & 21.56 & 19.72 & 22.99 & 1.07\\

Aya-command & 12.56 & 0.16 & 35.72 & 2.16 & 0.26 & 20.13 & 0.73 & 5.65 & 13.20 & 2.79 & 28.24 & 23.48 & 19.44 & 21.43 & 0.37\\
Claude-3 & 23.70 & 0.54 & \cellcolor{mediumgreen}70.02 & 5.64 & \cellcolor{lightblue}0.43 & \cellcolor{mediumgreen}45.57 & 1.09 & 40.40 & \cellcolor{lightblue}7.78 & \cellcolor{lightblue}1.32 & 28.64 & 19.02 & 31.19 & 22.91 & 0.88\\
Gemini-1.5 & 23.04 & 0.56 & 4.20 & 0.04 & 0.38 & 1.37 & 1.16 & 11.26 & 13.23 & 2.76 & 26.66 & 24.52 & 15.14 & 20.24 & 0.81\\

Few-shot GPT-3.5 & 22.82 & 0.88 & 54.14 & \cellcolor{lightblue}7.37 & 0.30 & 34.60 & 1.21 & 23.12 & 7.96 & 1.65 & 27.86 & 22.70 & 30.23 & 25.62 & 1.13\\
Zero-shot GPT-3.5 & 15.43 & 0.69 & 52.14 & 4.33 & 0.20 & 31.66 & 1.30 & 17.45 & 10.80 & 2.17 & 30.70 & 21.92 & 26.97 & 25.09 & 0.98\\
Fine-tune GPT-3.5 & 27.14 & 0.33 & 4.12 & 0.00 & 0.00 & 1.23 & 1.11 & \cellcolor{lightblue}71.66 & 7.36 & 1.46 & \cellcolor{lightblue}47.50 & \cellcolor{mediumgreen}44.58 & \cellcolor{lightblue}32.67 & \cellcolor{lightblue}28.64 & 1.08\\
CoT GPT-3.5 & 38.08 & 1.25 & 5.24 & 0.16 & 0.11 & 1.63 & 1.05 & 24.41 & 8.93 & 1.92 & 31.06 & 22.22 & 26.85 & 25.60 & 0.83\\

Few-shot GPT-4 & \cellcolor{lightblue}45.28 & 0.48 & 58.44 & 6.45 & 0.31 & 41.66 & 0.84 & 38.01 & 7.96 & 1.65 & 24.18 & 18.22 & 21.90 & 20.87 & 1.37\\
Zero-shot GPT-4 & 35.40 & 0.54 & 57.86 & 6.28 & 0.20 & 41.42 & 0.88 & 38.76 & 12.17 & 1.99 & 27.04 & 21.16 & 21.99 & 22.18 & 1.21\\

\midrule
\multicolumn{15}{c}{\textit{Open-Sourced Models (LLMs)}} \\
\midrule
\(\emojwen{\sc}{}\)Baichuan-13B & \cellcolor{lightblue}11.17 & 0.88 & \cellcolor{lightblue}33.20 & \cellcolor{lightblue}2.05 & 0.60 & \cellcolor{lightblue}22.62 & 1.20 & \cellcolor{lightblue}13.67 & 32.70 & 4.31 & 27.68 & 21.42 & 15.92 & 22.74 & 1.56\\
\(\emojwen{\sc}{}\)ChatGLM-6B & 10.30 & 0.68 & 6.94 & 0.50 & 0.00 & 6.33 & 1.35 & 1.38 & 29.68 & 4.25 & 26.88 & 12.60 & 12.43 & \cellcolor{lightblue}27.28 & 0.96\\
\(\emojwen{\sc}{}\)Chinese-LLaMA-7B & 5.13 & 0.97 & 9.26 & 0.64 & 0.17 & 6.32 & 1.92 & 0.32 & \cellcolor{lightblue}15.83 & \cellcolor{lightblue}3.00 & 26.26 & 24.86 & 13.42 & 22.32 & 0.93\\
\(\emojwen{\sc}{}\)InternLM-7B & 9.68 & 1.05 & 12.08 & 0.34 & 0.05 & 8.89 & 1.50 & 0.00 & 45.38 & 5.50 & \cellcolor{lightblue}28.82 & 24.66 & 13.38 & 22.01 & 0.95\\

\(\emojwen{\sc}{}\)Yi-6B & 8.86 & 0.70 & 14.18 & 1.05 & 0.21 & 12.14 & 1.40 & 0.32 & 29.49 & 4.24 & 28.56 & 22.40 & 7.76 & 24.17 & 0.85\\

Bloom-7B & 9.81 & 0.96 & 3.48 & 0.54 & 0.04 & 4.15 & 1.70 & 0.00 & 46.76 & 4.05 & 27.92 & \cellcolor{lightblue}24.96 & 14.47 & 23.19 & 0.87\\
Qwen-7B & 5.25 & 1.16 & 17.30 & 0.85 & 0.23 & 12.41 & 1.50 & 1.59 & 34.16 & 4.62 & 25.02 & 20.20 & \cellcolor{lightblue}21.92 & 23.30 & 1.30\\
Qwen-2-7B & 6.76 & 1.50 & 15.42 & 0.68 & 0.22 & 10.70 & 1.75 & 0.42 & 44.48 & 5.39 & 23.16 & 18.50 & 21.54 & 22.68 & 1.40\\
Orion-14B & 9.00 & 1.04 & 5.27 & 0.18 & \cellcolor{lightblue}0.76 & 9.46 & 1.11 & 3.39 & 31.45 & 4.45 & 28.40 & 22.82 & 19.38 & 24.81 & 0.90\\

\midrule

Fine-tune PIXEL & \cellcolor{mediumgreen} 84.57 & - & -& -& -& -& -& -& -& -& -& -& -& -& -\\
\cmidrule[1pt]{1-16}

Majority Baseline & 52.86 & - & 5.61 & 0.97 & 7.55 & 0.00 & - & 0.00 & 15.72 & 3.12 & 31.83 & 30.34 & 41.61 & 0.00 & -\\
\bottomrule
\end{tabular}
}
\caption{Model performance on Chinese character visuals on four different tasks (\S \ref{subsec:tasks}). \(\mathbf{H}\): Entropy, \(\emojwen{\sc}{}\): Chinese-English bilingual models. The top scores for each type of models (VLM/close-sourced LLM/open-sourced LLM) and all models are highlighted in \colorbox{lightblue}{blue} and \colorbox{mediumgreen}{green}, respectively. }
\label{tab:charTaskResult}
\end{table*}

\section{Evaluation on Recognizing Visual Information of Chinese Characters}
\label{sec:intrinsic}
To evaluate whether language models contain or can learn the visual information embedded in Chinese characters, we establish a benchmark by setting up a series of tasks (see example of each task in Figure \ref{fig:dataset}) derived from our dataset.
\subsection{Chinese Character Tasks}
\label{subsec:tasks}
\begin{CJK*}{UTF8}{gbsn}

\paragraph{Structure Recognition of Chinese Characters.}
We assess LLMs and VLMs' proficiency in identifying the correct structural arrangements of Chinese characters using a multiple-choice format. We present the character with eight struc0ture types and evaluate the model's answer using the F1 score.

\paragraph{Radical Recognition of Chinese Characters.}
We evaluate LLMs and VLMs' ability to recognize radical information in two tasks: character-to-radical and radical-to-character. For the first task, models receive a character and order guidelines, and are prompted to identify its radicals in sequence. Performance is measured by the accuracy of each radical in order and the overall F1 score. For the second task, models are given radicals and their relative positions and asked to identify the correct characters, with accuracy as the metric.

\end{CJK*}

\paragraph{Stroke Count Identification of Chinese Characters.}
We evaluate models' ability to identify the number of strokes required to write query characters with performance measured using Mean Absolute Error (MAE) and Mean Squared Error (MSE).

\paragraph{Stroke Decomposition of Chinese Characters.}
Similar to radical-to-character task, we evaluate LLMs and VLMs' ability to identify the sequence of strokes required to write the query character. Performance is measured by the accuracy of each stroke in order and the overall F1 score.

\subsection{Experimental Setup}
\label{subsec:llms-evaluation}
We assess the visual information of Chinese characters using multilingual, bilingual, and open-source LLMs and VLMs. Multilingual LLMs include Aya-command \cite{2024aya}, Claude-3 \cite{anthropic2024claude3}, Gemini-1.5, GPT-3.5 Turbo \cite{openai2024gpt35turbo}, and GPT-4 \cite{openai2023gpt4}. Chinese-English bilingual LLMs include ERNIE-Lite \cite{baidu2024ernie35}, Kimi-v1 \cite{moonshot2024kimi}, and open-source LLMs such as Baichuan-13B \cite{baichuan2024baichuan13b}, BLOOM-7B \cite{bigscience2024bloom7b1}, ChatGLM-6B \cite{zeng2023glm130b}, Chinese-LLaMA-7B \cite{hfl2024chinesellama2}, InternLM-7B \cite{internlm2024internlm7b}, Orion-14B \cite{chen2024orion14b}, Qwen-7B \cite{qwen}, Qwen-2-72B, and Yi-6B \cite{ai2024yi}.
We also evaluate VLMs by providing images of query characters in the Microsoft YaHei\footnote{YaHei is the default Chinese font in Microsoft Office.} font to vision-capable models, including Claude-3, Gemini-1.5, GPT-4, GPT-4o \cite{openai_gpt4o_system_card_2024}, and bilingual models Ernie-4 \cite{baidu2024yiyan} and Kimi-v1. Additionally, we assess the pixel-based encoder model, PIXEL \cite{rust-etal-2023-pixel}. Since PIXEL is limited to specific tasks such as span-based QA, it is evaluated only on the multiple-choice structure recognition task after fine-tuning. To explore the models' ability to learn Chinese visual information, we apply Chain-of-Thought (CoT) prompting and fine-tuning to GPT-3.5, and few-shot settings to both GPT-3.5 and GPT-4. The remaining models are evaluated in a zero-shot setting. We repeat the evaluation for each task five times to compute entropy, using it as an indicator of the models' confidence. Detailed prompting and fine-tuning procedures are provided in Appendix \ref{subsec:settingdetail}.

\subsection{Experimental Results}

\begin{CJK*}{UTF8}{gbsn}
As shown in Table \ref{tab:charTaskResult}, the majority of models demonstrate only a vague understanding across Chinese character-visual tasks. Among the evaluated models, Chinese-English bilingual VLMs achieve the highest overall performance, effectively leveraging visual information from the images. Multilingual VLMs, however, perform similarly to their LLM counterparts, with both groups achieving better-than-random-guess results. The performance of vision-lacking LLMs suggests that they have likely encountered textual data discussing knowledge about radicals during pre-training as shown in Appendix \ref{subsec:o1and4o}. In contrast, open-source LLMs often perform worse than random guesses.

\paragraph{Structure Recognition Task.}
Most models tend to struggle with the structural arrangement of Chinese characters, with F1 scores below 50\%. A notable exception is PIXEL, which achieves an outstanding F1 score of 84.57. Despite being pre-trained solely on an English corpus (English Wikipedia and BookCorpus) and exposed to Chinese only during fine-tuning, PIXEL \cite{razzhigaev-etal-2022-pixel} demonstrates strong potential for Chinese language processing by naturally capturing visually embedded information. GPT-3.5 also saw a 75\% performance increase from zero-shot to fine-tuning settings. To better understand the performance boost after applying learning methods, we further examine the impact of Chinese character encoding and potential mapping patterns between a character's structure and Unicode in Appendix \ref{subsec:encoding}.

\paragraph{Radical Recognition Task.}
In the character-to-radical task, a clear trend emerges where model performance is the highest for the first component and sharply decreases for subsequent ones. For example, Claude-3 achieves an F1 score of 70.02 for the first component, but this drop to 5.64 for the second component and nearly zero for the third. This pattern suggests that models could possibly associate the meaning of the radical with the character, as the first radical often relates to the semantic attribute of the character, such as ``艹'' in ``花.'' Interestingly, fine-tuning, CoT prompting, and the addition of vision in multilingual models drastically decreased performance of character-to-radical task to nearly zero. However, in the radical-to-character task, fine-tuning GPT-3.5 results in a significant improvement, achieving an F1 score of 71.66. The reason for this disparity, aside from the inherent difficulties between the two tasks, could be that fine-tuning and query characters for the radical-to-character task come from a subset of more common characters, for which we annotated structure information. In contrast, the character-to-radical task includes more complex and rarely used characters, potentially leading to catastrophic learning failure.
\paragraph{Stroke Decomposition and Count Tasks.}
Overall, most models struggle with identifying specific stroke compositions, but demonstrate a general understanding of stroke count. For instance, Claude-3 achieves the lowest MSE among all LLMs at 7.78, significantly lower than the dataset's average stroke count of 11.51. Both tasks benefit from the applied learning methods. For stroke composition, fine-tuning yields the best results, while for stroke count, all methods show similar levels of improvement.

\begin{figure}[t]
    \centering
    \includegraphics[width=1\linewidth]{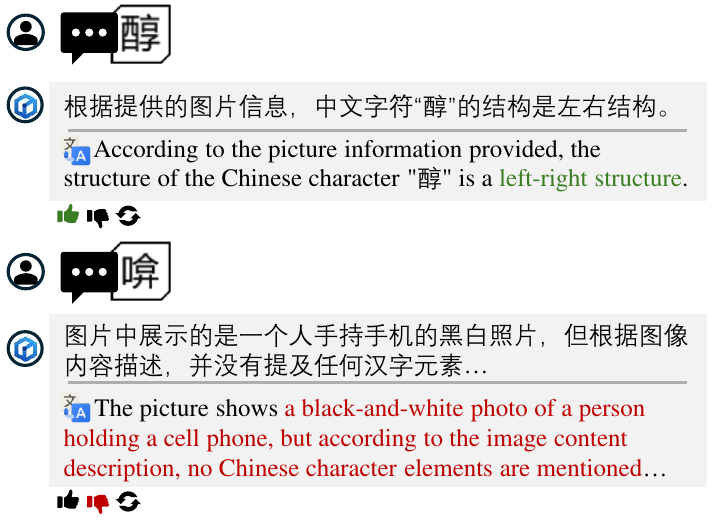}
    \caption{Example responses generated by Ernie-4 with vision, translated using Google Translate.}
    \label{fig:errorErnie}
\end{figure}
\subsection{Error Analysis of Bilingual VLMs}
To better understand the superior performance of bilingual VLMs, we conducted an error analysis on Ernie-4 and Kimi-v1 with vision. Both models exhibit common patterns of mistakes across several types of characters. First, complex and dense characters are often misrecognized as some other more frequently used characters that look similar. As the complexity of characters increases, the individual radicals become more compressed within the available space, which can lead to misrecognition between similar characters, such as ``黄'' and ``\includegraphics[scale=0.09]{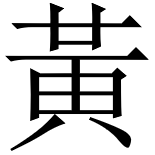}''. Second, characters with only a single stroke difference are frequently mistaken for their more common variant. Third, when dealing with rare characters, Ernie-4 often fails to detect any character in the image, while Kimi-v1 may refuse to allow the user to send the prompt if it cannot extract the character. Additionally, both models sometimes mistake one radical for the entire character or confuse characters with black-and-white photos as shown in Figure \ref{fig:errorErnie}. More examples of Kimi-v1 and Ernie-4 are provided in Appendix \ref{subsec:ernieeg}.
\end{CJK*}

\begin{table}[t!]
\centering
\setlength{\tabcolsep}{3.5pt}
\small
\begin{tabular}{l c c c c c c c}
\toprule
\multirow{3}{*}{\textbf{Models}}& \multicolumn{6}{c}{\textbf{Name Entity Recognition}}\\
\cmidrule(lr){2-7} 
 & \multicolumn{3}{c}{\textbf{People's Daily}} & \multicolumn{3}{c}{\textbf{Weibo}}   \\

\cmidrule(lr){2-4} \cmidrule(lr){5-7} 
 & \textbf{B} & \multicolumn{2}{c}{\textbf{RP}} & \textbf{B} & \multicolumn{2}{c}{\textbf{RP}} \\
\midrule

Aya        & 52.00 & 54.61 & \cellcolor{stronggreen}(+2.6) & 24.78 & 16.00 & \cellcolor{red!65}(-8.8)\\
Claude-3   & 68.54 & 70.48 & \cellcolor{lightgreen}(+1.9) & 41.08 & 41.67 & \cellcolor{mediumgreen}(+1.6)\\
ERNIE-Lite & 7.55  & 21.05 & \cellcolor{extremedarkgreen}(+14) & 6.25 & 14.81 & \cellcolor{verydarkgreen}(+8.6)\\
GPT-3.5    & 55.74 & 55.96 & \cellcolor{lightgreen}(+0.2) & 38.37 & 44.87 & \cellcolor{extremedarkgreen}(+12)\\
GPT-4      & 65.23 & 65.96 & \cellcolor{mediumgreen}(+0.7) & 38.59 & 40.34 & \cellcolor{mediumgreen}(+1.8)\\
QWen 72B   & 58.81 & 58.94 & \cellcolor{lightgreen}(+0.1) & 29.39 & 33.17 & \cellcolor{stronggreen}(+3.8)\\

\bottomrule
\end{tabular}
\caption{Model performances for NER evaluated solely on samples where the model identifies unfamiliar words.}
\label{tab:ner2}
\end{table}

\section{Evaluation on Utilizing Radicals}
\label{sec:radicalPrompting}
\begin{figure}[t!]
    \centering
    \includegraphics[width=1\linewidth]{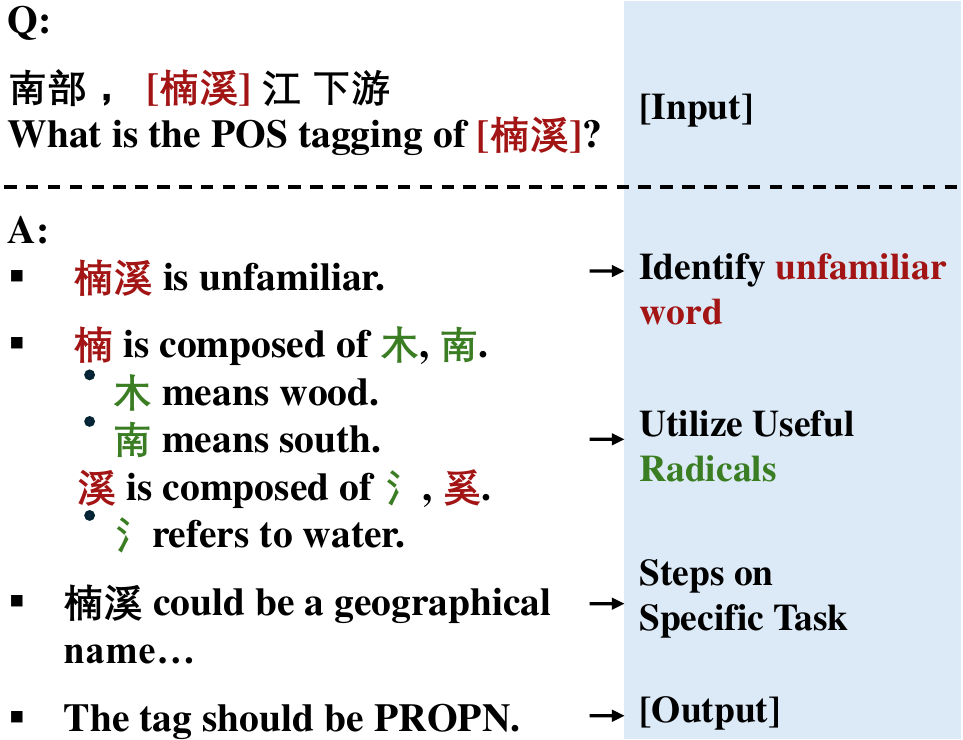}
    \caption{Example of model's answers for part-of-speech (POS) tagging with an unfamiliar Chinese word using radical prompting.}
    \label{fig:promptingexample}
\end{figure}

We evaluated LLMs on downstream tasks, specifically examining performance differences when models are prompted vs. not prompted to use their knowledge of radicals to infer the meaning of unfamiliar words. Example is shown in Figure \ref{fig:promptingexample}.

\subsection{Downstream NLP Tasks}
Although LLMs may not match supervised LMs in traditional tasks, we chose these as key indicators of Chinese understanding to track improvements when models utilize radical information.

\paragraph{Part-of-Speech (POS) tagging.} 
For the POS tagging task, we selected a 5-word window containing at most one punctuation mark and tasked the model with identifying the POS tag of the central word. The model’s performance was evaluated using the F1 score. 
\begin{CJK*}{UTF8}{gbsn}
To cover a diverse range of sentences, we utilized three datasets: the GSD Simplified dataset \cite{gsdsimp2023}, the Parallel Universal Dependencies (PUD) dataset \cite{pud2023pos}, and a new dataset of 500 sentences from Tang Dynasty poems (written between 618 and 907 CE), processed using Classical Chinese RoBERTa \cite{robertaClassicalChinese2023}. We created and annotated this poetry dataset to evaluate how well radicals perform on Classical Chinese (文言), which is characterized by compact and precise language where more information is carried by each character.
Additionally, we conducted an ablation study with varied input window sizes, detailed in Appendix \ref{subsec:windowsize}.
\end{CJK*}

\paragraph{Named Entity Recognition (NER).} 

We tasked the model with identifying three types of entities — PER (person), LOC (location), and ORG (organization) — at the character level, using the BIO tagging standard. We excluded nominal entities provided in some datasets to streamline the analysis. The model’s performance was evaluated using the F1 score. We use two distinct datasets for the NER task: the People's Daily dataset \cite{renmindailynerdataset2023}, which focuses on formal Chinese text, and the Weibo NER dataset \cite{pengdredze2015weiboner}, which is oriented towards casual and online text.

\paragraph{Chinese Word Segmentation (CWS).} CWS is a unique task in Chinese language processing. Distinguished from many other languages, Chinese does not use delimiters such as white spaces to separate words within sentences. Accurately segmenting text could be beneficial for many CLP applications. For this task, we provide sentences from the GSD and PUD and ask models to segment them into words. Answers are evaluated using the F1 score.

\begin{table*}[t]
\centering
\setlength{\tabcolsep}{3.5pt}
\fontsize{12}{1}
\resizebox{\textwidth}{!}{
\begin{tabular}{l c c@{\hskip 5pt}p{26pt} c@{\hskip 5pt}p{26pt} c c@{\hskip 5pt}c c@{\hskip 5pt}c c c@{\hskip 5pt}c c@{\hskip 5pt}p{26pt}}
\toprule
\multirow{3}{*}{\textbf{Models}} & \multicolumn{15}{c}{\textbf{Part-Of-Speech Tagging}} \\
\cmidrule(lr){2-16}
& \multicolumn{5}{c}{\textbf{GSD}} & \multicolumn{5}{c}{\textbf{PUD}} & \multicolumn{5}{c}{\textbf{Poems}} \\
\cmidrule(lr){2-6} \cmidrule(lr){7-11} \cmidrule(lr){12-16}
 & \textbf{B} & \multicolumn{2}{c}{\textbf{RP}} &  \multicolumn{2}{c}{\textbf{RP (Oracle)}} & \textbf{B} & \multicolumn{2}{c}{\textbf{RP}} & \multicolumn{2}{c}{\textbf{RP (Oracle)}} & \textbf{B} & \multicolumn{2}{c}{\textbf{RP}} & \multicolumn{2}{c}{\textbf{RP (Oracle)}} \\
\midrule
Aya-command & 68.86 & 68.91 & \cellcolor{lightgreen}(+0.1) & 70.41 & \cellcolor{mediumgreen}(+1.6) & 73.87 & 77.21 & \cellcolor{stronggreen}(+3.3) & 76.95 & \cellcolor{stronggreen}(+3.1) & 64.71 & 64.72 & \cellcolor{lightgreen}(+0.0) & 65.54 & \cellcolor{lightgreen}(+0.8) \\
Claude-3 & 69.37 & 70.68 & \cellcolor{mediumgreen}(+1.3) & 70.45 & \cellcolor{mediumgreen}(+1.1) & 69.37 & 70.45 & \cellcolor{mediumgreen}(+1.1) & 70.68 & \cellcolor{mediumgreen}(+1.3) & 65.53 & 66.20 & \cellcolor{lightgreen}(+0.7) & 66.71 & \cellcolor{mediumgreen}(+1.2) \\
ERNIE-Lite-8K & 27.06 & 24.97 & \cellcolor{red!65}(-2.1) & 32.73 & \cellcolor{verydarkgreen}(+5.7) & 30.35 & 30.29 & \cellcolor{red!20}(-0.0) & 41.29 & \cellcolor{extremedarkgreen}(+10.9) & 44.19 & 42.17 & \cellcolor{red!65}(-2.0) & 49.07 & \cellcolor{stronggreen}(+4.9) \\
GPT-3.5 & 59.08 & 64.62 & \cellcolor{verydarkgreen}(+5.5) & 67.56 & \cellcolor{verydarkgreen}(+8.5) & 62.61 & 69.90 & \cellcolor{verydarkgreen}(+7.3) & 73.46 & \cellcolor{extremedarkgreen}(+10.9) & 53.51 & 59.22 & \cellcolor{verydarkgreen}(+5.7) & 61.39 & \cellcolor{verydarkgreen}(+7.9) \\
GPT-4 & 71.55 & 72.14 & \cellcolor{lightgreen}(+0.6) & 72.95 & \cellcolor{mediumgreen}(+1.4) & 76.20 & 76.72 & \cellcolor{lightgreen}(+0.5) & 77.35 & \cellcolor{mediumgreen}(+1.2) & 66.94 & 67.11 & \cellcolor{lightgreen}(+0.2) & 67.57 & \cellcolor{lightgreen}(+0.6) \\
o1-mini & 63.24 & 67.96 & \cellcolor{stronggreen}(+4.7) & 64.31 & \cellcolor{mediumgreen}(+1.1) & 70.37 & 71.42 & \cellcolor{mediumgreen}(+1.1) & 75.49 & \cellcolor{stronggreen}(+4.1) & 47.73 & 50.04 & \cellcolor{mediumgreen}(+2.3) & 49.00 & \cellcolor{mediumgreen}(+1.3) \\
QWen-72B & 62.20 & 65.38 & \cellcolor{stronggreen}(+3.2) & 67.32 & \cellcolor{verydarkgreen}(+5.1) & 60.09 & 64.70 & \cellcolor{stronggreen}(+4.6) & 66.90 & \cellcolor{verydarkgreen}(+6.8) & 55.63 & 57.78 & \cellcolor{mediumgreen}(+2.2) & 59.54 & \cellcolor{stronggreen}(+3.9) \\

\bottomrule
\end{tabular}

}
\caption{Model performances for POS tagging with baseline (B), radical prompting without golden components (RP), and radical prompting with oracle information (RP (Oracle)). Performance change relative to baseline is highlighted with \textcolor{darkgreen}{green} for increase and \textcolor{red}{red} for decrease.}
\label{tab:pos}
\end{table*}

\begin{table*}[t]
\centering
\setlength{\tabcolsep}{3.5pt}
\small
\begin{tabular}{l c c c c c c c c c c c c c c}
\toprule
\multirow{3}{*}{\textbf{Models }}& \multicolumn{6}{c}{\textbf{Name Entity Recognition}}& & \multicolumn{6}{c}{\textbf{Chinese Word Segmentation}}\\
\cmidrule(lr){2-7} \cmidrule(lr){9-14}
 & \multicolumn{3}{c}{\textbf{People's Daily}} & \multicolumn{3}{c}{\textbf{Weibo}} & & \multicolumn{3}{c}{\textbf{GSD}} & \multicolumn{3}{c}{\textbf{PUD}}  \\

\cmidrule(lr){2-4} \cmidrule(lr){5-7}  \cmidrule(lr){9-11} \cmidrule(lr){12-14} 
 & \textbf{B} & \multicolumn{2}{c}{\textbf{RP}} & \textbf{B} & \multicolumn{2}{c}{\textbf{RP}} && \textbf{B} & \multicolumn{2}{c}{\textbf{RP}} & \textbf{B} & \multicolumn{2}{c}{\textbf{RP}}\\
\midrule

Aya-command & 38.24 & 36.36 & \cellcolor{red!50}(-1.88) & 37.88 & 30.83 & \cellcolor{red!65}(-7.05) & & 87.98 & 89.08 & \cellcolor{stronggreen}(+1.10) & 88.68 & 91.05 & \cellcolor{verydarkgreen}(+2.37) \\
Claude-3 & 69.74 & 73.79 & \cellcolor{verydarkgreen}(+4.05) & 45.64 & 46.86 & \cellcolor{mediumgreen}(+1.22) & & 94.90 & 95.16 & \cellcolor{lightgreen}(+0.26) & 94.12 & 94.96 & \cellcolor{lightgreen}(+0.84) \\
ERNIE-Lite-8K & 12.10 & 12.99 & \cellcolor{mediumgreen}(+0.89) & 6.72 & 6.90 & \cellcolor{lightgreen}(+0.19) & & 88.04 & 88.70 & \cellcolor{mediumgreen}(+0.66) & 69.54 & 73.57 & \cellcolor{verydarkgreen}(+4.03) \\
GPT-3.5 & 56.89 & 55.97 & \cellcolor{red!20}(-0.92) & 36.65 & 36.64 & \cellcolor{red!20}(-0.01) & & 95.68 & 94.87 & \cellcolor{red!20}(-0.81) & 93.91 & 93.70 & \cellcolor{red!20}(-0.21) \\
GPT-4 & 66.04 & 68.05 & \cellcolor{stronggreen}(+2.01) & 43.83 & 44.68 & \cellcolor{mediumgreen}(+0.85) & & 94.21 & 94.88 & \cellcolor{mediumgreen}(+0.67) & 94.24 & 95.63 & \cellcolor{stronggreen}(+1.39) \\
o1-mini & 84.21 & 91.67 & \cellcolor{verydarkgreen}(+7.46) & 56.37 & 69.70 & \cellcolor{extremedarkgreen}(+13.3) & &  97.21 & 100.0 & \cellcolor{verydarkgreen}(+2.79) & 93.65 & 97.00 & \cellcolor{verydarkgreen}(+3.35)\\
QWen 72B & 62.73 & 59.59 & \cellcolor{red!65}(-3.14) & 31.78 & 35.83 & \cellcolor{verydarkgreen}(+4.05) & & 96.59 & 95.57 & \cellcolor{red!50}(-1.02) & 89.79 & 91.94 & \cellcolor{verydarkgreen}(+2.15) \\

\bottomrule
\end{tabular}
\caption{Model performances for NER and CWS tasks with baseline (B) and radical prompting (RP).}
\label{tab:ner}
\end{table*}

\subsection{Experimental Setup}
We select a series of LLMs for evaluation, including Aya-command, Claude-3, ERNIE-Lite-8K, GPT-3.5, GPT-4, and QWen-1.5 72B Chat. The models are instructed to return answers in JSON format, with target sentences annotated in a manner similar to \citet{blevins-etal-2023-prompting}. Each task and dataset is evaluated using 2,000 sample sentences five times. Due to higher costs, Claude-3 and GPT-4 are evaluated with 1,000 samples. Additionally, we experiment with o1-mini \cite{openai_o1_mini_2024} using 100 samples, repeated three times. We experiment with three different prompting methods:

\paragraph{Baseline.} Our baseline employs the CoT prompting framework with steps that guide the model to execute tasks. See prompts in Appendix \ref{subsec:basepromptinglines}. 

\paragraph{Radical Prompting.}
We incorporate the radical information into the input prompt as steps within the CoT framework. The process begins with the model identifying any unclear words within a given context. Then, the model is instructed to dissect these words into their constituent radicals and attempt to utilize useful radicals to aid the task. Steps are then provided to guide the model in executing specific tasks, identical to the baseline, with three examples. When using radical prompting, it is important to guide models to critically assess information from character components to avoid being misguided. Therefore, one of the three examples intentionally includes radical that can be misleading, helping the model learn to discern when to use radical information. Prompt lines of radical prompting are listed in Appendix \ref{subsec:promptinglines}.

\paragraph{Radical Prompting (Oracle).}
Similar to the radical prompting method, instead of instructing the model to decompose characters, we directly provided the correct radicals in the input prompt. This method was applied only to the POS tagging task, as it required supplying the radical of just the central word. For the other tasks, it is impractical to provide radicals for all characters in the sentence.

\subsection{Experimental Results}
Our results suggest that radicals have strong potential if models can properly understand and utilize them. Results for POS tagging are shown in Table \ref{tab:pos} with qualitative analysis in Appendix \ref{subsec: ic}. In the POS tagging task, models consistently show improvement across datasets, especially when the correct radicals are provided. Notably, in the PUD dataset, ERNIE-Lite-8K exhibits a slight decrease in performance without the correct radicals but shows an increase of approximately 11 F1 points when the correct radicals are included. 

For the NER and CWS tasks, the initial results are mixed as shown in Table \ref{tab:ner}. However, our error analysis reveals that with the radical prompting method, incorrect answers often occur when the model bypasses the use of radicals and asserts that there are no ambiguous words in the sentence being examined. This suggests that the negative effect may be attributed to the longer prompts, as more robust models, such as Claude-3 and GPT-4, still demonstrate improvement in performance across datasets. When evaluating only the samples where the model identifies ambiguous words in the radical prompt setting, we find that the models genuinely perform better, as shown in Table \ref{tab:ner2}. However, a notable exception is Aya-command, whose performance drops significantly on the Weibo dataset. Upon closer examination, we find that Aya has a strong tendency to split words into individual characters rather than into radicals. Example of such output is shown in Appendix \ref{ayaeg}.


\section{Conclusion}
In this paper, we create a comprehensive benchmark on visual information embedded in Chinese characters. Our evaluation of the benchmark highlight the suboptimal performance of LLMs and VLMs in handling information below the character level. Despite this, our experiments with \textit{radical prompting} demonstrate that these sub-character features can still be beneficial. The results show consistent improvements in POS tagging when correct radicals are provided, and promising results in NER on sentence contains unfamiliar words. Our work highlights the potential of radical knowledge for Chinese NLP applications and advocates attention to help models leverage it, including additional training on radicals or improving Chinese digital systems to more effectively integrate radical structures.

\section*{Limitations}
While our study offers valuable insights into the integration of radical prompting in Chinese language models, it also highlights areas for further exploration. First, the dataset used in this research does not represent the full range of Chinese characters, as the majority are sourced from simplified Chinese.

Moreover, the study primarily evaluates radical prompting on a limited selection of models and tasks, which may not fully capture its potential across a wider range of models and language processing applications.

Lastly, an area for improvement in our methodology involves the exclusive use of English in our prompt lines. Incorporating Chinese in the prompting strategy could further enhance the relevance and effectiveness of the prompts, better aligning with the linguistic context of the target language.

\section*{Acknowledgements}
We extend our gratitude to Dengchun Yuan for his inspiring discussions on this work and to Yao Dao for offering valuable advice on prompting design. We also thank Duong Minh Le for his helpful guidance on writing. Finally, we appreciate the efforts of Geyang Guo, Shu Zhu, and Wei Zhou for their assistance with the validation process.

\bibliography{anthology,custom}

\appendix

\section{General Experiment Details}
\paragraph{Model Snapshots}
The experiments incorporated different versions of widely recognized models to evaluate their performance in processing Chinese characters. The specific snapshots used for each model are as follows:
\begin{itemize}
\item \textbf{GPT-3.5} and \textbf{GPT-4} were used with the snapshot dated \textit{2023-11-06}.
\item \textbf{Claude} model's evaluation utilized the \textit{2024-02-29} snapshot.
\item \textbf{Ernie-Lite-8K} was tested using the \textit{2023-09-22} snapshot.
\end{itemize}

\paragraph{Temperature Settings}
\begin{itemize}
\item \textbf{Aya}, \textbf{Yi-6B}, \textbf{Qwen-7B-Chat}, \textbf{Baichuan-13B}, and \textbf{Mistral-7B} were set at a lower temperature of \textit{0.3} as recommended.
\item For \textbf{other models} not specifically mentioned, a temperature setting of \textit{0.7} was used.
\end{itemize}
\section{Validation on Radical Annotation}
\label{sec:annotation}
To ensure the quality of our radical annotations, we conducted a validation process with a team of four annotators. This team included three volunteer native Chinese speakers and one native Chinese-speaking author. Prior to the formal annotation process, each annotator underwent a brief practice session to familiarize themselves with the task. Following this, the speed of checking 100 rows is around two and a half minutes per annotator.

For the validation task, we randomly assigned 3,500 rows to each annotator from our dataset, ensuring that the task could be completed within roughly two hours. Each annotator was instructed to review the assigned rows and flag any errors they identified in the radical annotations. The errors flagged were then collectively analyzed to compute the error rate and guide the necessary revisions.
\section{Details on Visual Info Evaluation}

\subsection{Detail Settings}
\label{subsec:settingdetail}
For our evaluation, we use different sampling methods and settings based on the type of model. For LLMs, a random sample of 1,000 characters is selected for each task and model. Due to higher costs, the number of samples for VLMs is reduced to 500. ERNIE-V and Kimi-V, which lack API access, are tested manually with only 100 samples. We incorporate few-shot learning by providing models with three examples for each task, except for the structure recognition task, where one example per structure type is given. In the Chain-of-Thought (CoT) setting, models are prompted to break down their reasoning process step-by-step, with detailed prompts provided in the Appendix \ref{subsec:cotprompt}. Models with fine-tuning are trained in OpenAI's platform using cross-entropy loss with a 7:3 split and tested using 1,000 samples randomly selected from the CCD dataset. To assess consistency and model entropy, each question is asked five times, and the best trial out of the five for each task is selected to calculate the overall results.

To adapt answers from models generating long responses conventionally, we first let models generate responses freely without a specific answer format. Then, we use GPT-3.5 Turbo to extract answers from various model responses. For open-source models and extraction-used GPT-3.5 Turbo, a temperature of 0.3 is applied. Closed-source models generally use a temperature of 0.7 unless otherwise recommended by model documentation.

\subsection{CoT Prompting}
\label{subsec:cotprompt}
\begin{figure*}
    \centering
    \includegraphics[width=1\linewidth]{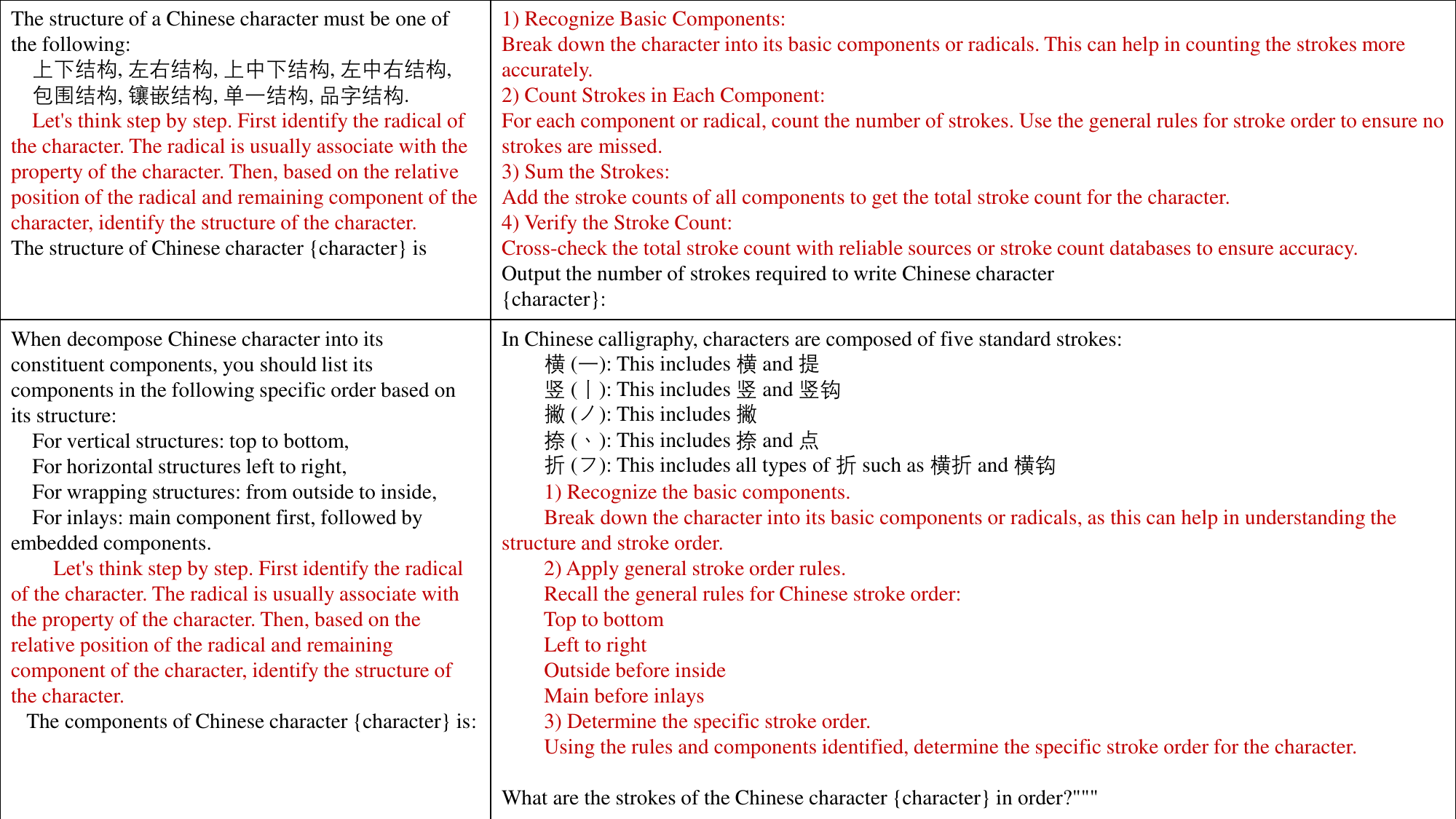}
    \caption{Prompt Line of Visual Tasks. Sub-windows are organized as follows: Structure (Top Left), Radical (Bottom Left), Stroke Count (Top Right), and Stroke Decomposition (Bottom Right). The red section indicates the Chain-of-Thought (CoT) prompting portion.}
    \label{fig:bishunprompt}
\end{figure*}

We present the prompt lines used for visual info evaluation in Figure \ref{fig:bishunprompt}.



\subsection{Chinese VLMs Behavior}
\begin{figure*}
    \centering
    \includegraphics[width=1\linewidth]{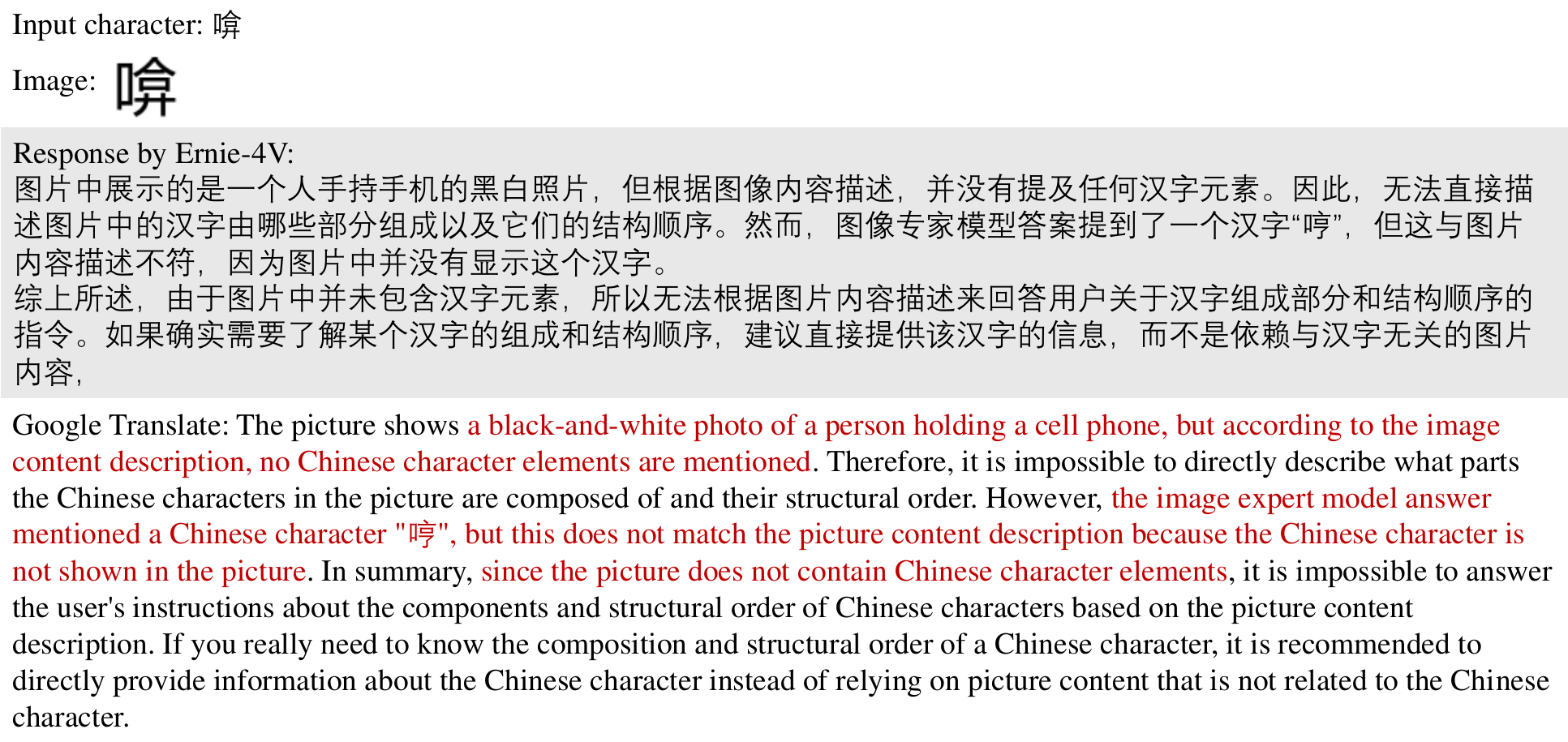}
    \caption{Example of Ernie-4 with vision response to rare character with English translation.}
    \label{fig:rarechareg}
\end{figure*}

\begin{figure*}
    \centering
    \includegraphics[width=1\linewidth]{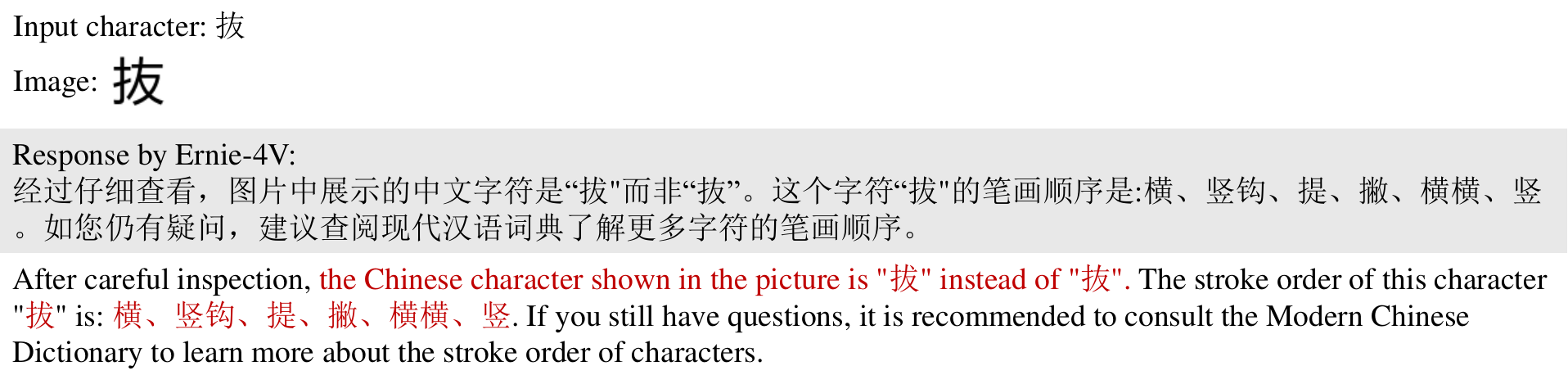}
    \caption{Example of Ernie-4 with vision response to extremely similar character with English translation.}
    \label{fig:simichareg}
\end{figure*}

\begin{figure*}
    \centering
    \includegraphics[width=1\linewidth]{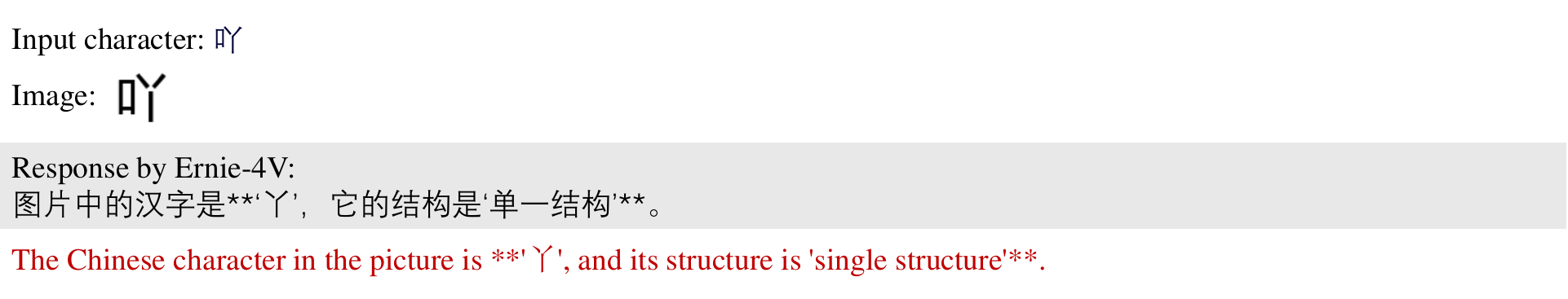}
    \caption{Example of Ernie-4 with vision response to part of the character as answer with English translation.}
    \label{fig:partchareg}
\end{figure*}

\begin{figure*}
    \centering
    \includegraphics[width=1\linewidth]{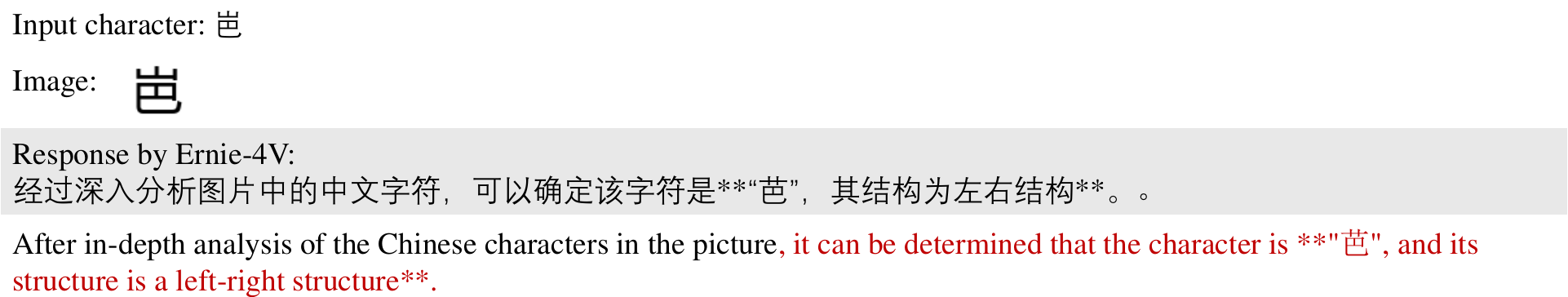}
    \caption{Example of Ernie-4 with vision response a character with different component part as answer with English translation.}
    \label{fig:radchareg}
\end{figure*}

\begin{figure*}
    \centering
    \includegraphics[width=1\linewidth]{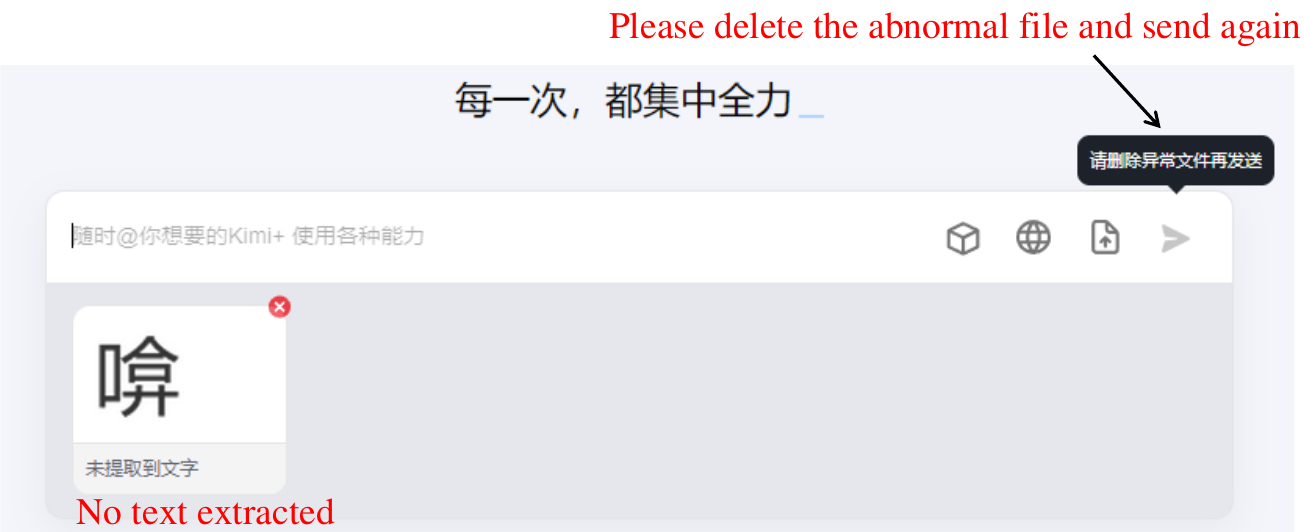}
    \caption{Example of Kimi-v1 with vision reject rarely used character with English translation.}
    \label{fig:errchareg}
\end{figure*}

Examples of VLMs misrecognizing images are shown in Figure \ref{fig:rarechareg}, \ref{fig:simichareg}, \ref{fig:partchareg},  \ref{fig:radchareg}, and \ref{fig:errchareg}.
\label{subsec:ernieeg}

\begin{CJK*}{UTF8}{gbsn}
\begin{table*}
\centering
\begin{tabular}{lll|lll}

\toprule
\textbf{Unicode} & \textbf{Character} & \textbf{Structure} & \textbf{Unicode}  & \textbf{Character} & \textbf{Structure}\\
\midrule

U+4EBF & 亿 & Left-Right &  U+4ED9 &  仙 & Left-Right\\
U+4EC0 & 什 & Left-Right &  U+4EE3 &  代 & Left-Right\\
U+4EC1 & 仁 & Left-Right &  U+4EEA &  仪 & Left-Right\\
U+4EC3 & 仃 & Left-Right &  U+4EEB &  仫 & Left-Right\\
U+4EC4 & 仄 & Wrapping &  U+4EF0 &  仰 & Left-Right\\
U+4EC7 & 仇 & Left-Right &  U+4EF2 &  仲 & Left-Right\\
U+4ECE & 从 & Left-Right &  U+4EF5 &  仵 & Left-Right\\
U+4ED1 & 仑 & Top-Bottom &  U+4EFB &  任 & Left-Right\\
U+4ED3 & 仓 & Top-Bottom &  U+4EFD &  份 & Left-Right\\
U+4ED5 & 仕 & Left-Right &  U+4F01 &  企 & Top-Bottom\\
U+4ED6 & 他 & Left-Right &  U+4F0A &  伊 & Left-Right\\
U+4ED7 & 仗 & Left-Right &  U+4F0D &  伍 & Left-Right\\
U+4ED8 & 付 & Left-Right &  U+4F0E &  伎 & Left-Right\\
\midrule

\end{tabular}

\caption{This table showcases a randomly selected range of Unicode characters in dataset along with their respective structures. This representation provides a snapshot of the structural information inherent in the Unicode.}
\label{tab:uni}
\end{table*}
\end{CJK*}

\subsection{GPT-4o and GPT-o1 Behavior}
\label{subsec:o1and4o}
\begin{figure*}
    \centering
    \includegraphics[width=1\linewidth]{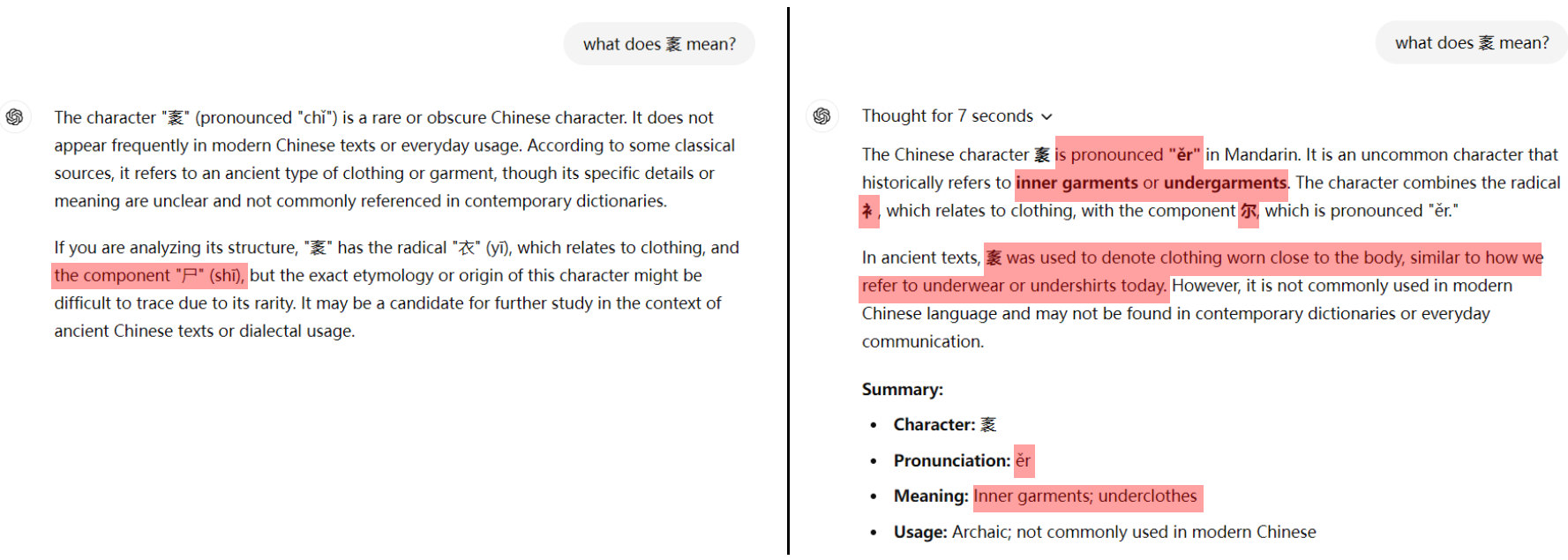}
    \caption{An example of GPT-4o and o1-preview's response, with incorrect statements highlighted in \color{red}{red}.}
    \label{fig:err4o}
\end{figure*}

When simply prompting GPT-4o and GPT-4.1 with the question "What does this character mean?" for unfamiliar characters, both models provide responses with radical information, but they do so incorrectly (as shown in Figures \ref{fig:err4o}). This suggests that while the models have been exposed to radicals in their training corpus, they are still unable to correctly identify or interpret the appropriate radicals.
\section{Analysis on Chinese Character Encoding}
\label{subsec:encoding}
\subsection{Experiment on Encoding}
To further investigate why models after fine-tuning perform exceptionally well on structure tasks but show decreased performance on other Chinese visual tasks, we conducted a side experiment on different encoding systems to determine if they learn some sort of implicit pattern from the encoding.

\paragraph{Setup.}
We fine-tuned GPT-3.5 by explicitly switching all Chinese characters in the training and testing documents to various encodings—namely, Unicode, stroke, Pinyin\footnote{Pinyin is the Romanization of the Chinese characters based on their pronunciation. In Mandarin, it's the standard method for typing Chinese characters.}, Wubi, and Cangjie\footnote{Wubi and Cangjie are two glyph-based input methods that are uncommon to use.}—and evaluated them on the structure recognition task to assess the impact of these representations on the model's learning ability with visual knowledge of Chinese characters.

\begin{table}
\centering
\normalsize
\fontsize{9}{9}

\begin{tabular}{ll}

\toprule
\textbf{Encoding} & \textbf{Structure Acc}\\
\midrule

Unicode  &  39.80\\
Stroke  & 43.80\\
PinYin  & 13.85\\
WuBi & 11.81\\
CangJie & 11.66\\

\bottomrule
\end{tabular}

\caption{GPT-3.5 fine-tuning' performance on different way of encoding.}
\label{tab:embeddings}
\end{table}

\paragraph{Results.}
The results shown in Table \ref{tab:embeddings} indicate that Unicode encoding performs comparably to the vision-rich stroke encoding and significantly outperforms Pinyin encoding, which is limited to phonetic information. Upon further investigation, we found that the order of Chinese characters in Unicode is closely related to the stroke count and structure of the characters: Unicode is ordered by the stroke count of their indexing radical and the stroke count of remaining parts. However, the full potential of Unicode is diminished by numerous exceptions and a broad spectrum of extensions that complicate its utility in conveying visual knowledge, where similar structures are likely grouped together with stroke counts in incremental order, as detailed in Table \ref{tab:uni}. 

\begin{figure}[h]
    \centering
    \includegraphics[width=1\linewidth]{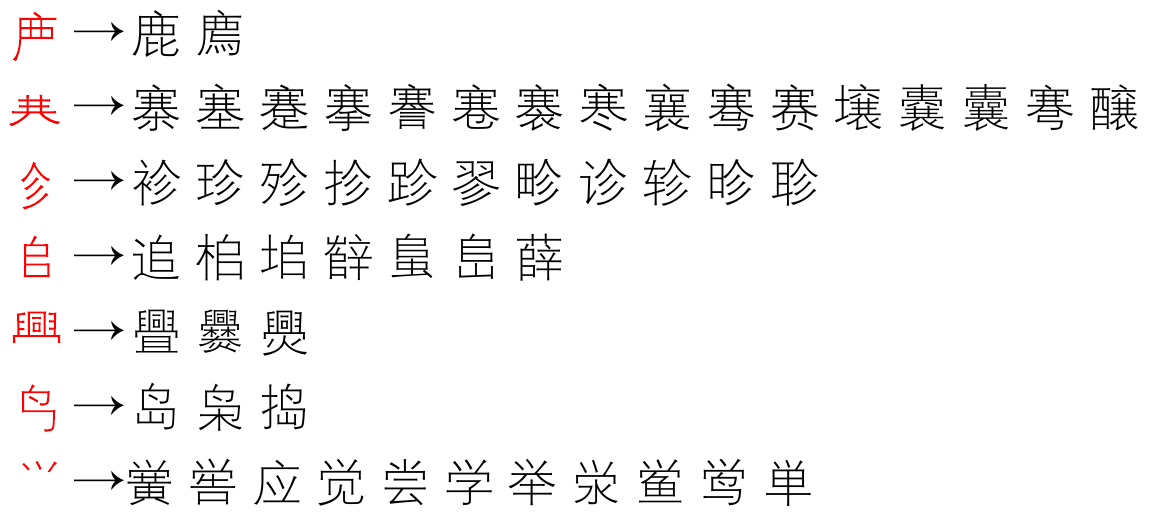}
    \caption{Examples of untypable radicals}
    \label{fig:eg-untypable}
\end{figure}

\subsection{Challenges on Radical Encoding}
\label{subsec:discussencoding}
None of the encoding systems mentioned above can fully exploit the potential of radicals. Stroke-based systems over-decompose characters into individual strokes, losing meaningful structure, while glyph-based input methods like Wubi and Cangjie oversimplify and over-categorize characters to prioritize efficiency as input methods.

However, a significant challenge lies in developing a radical-based encoding system. While some radicals have corresponding Unicode representations, they cannot be typed using standard input methods. With around 100,000 CJK ideographs in Unicode, the task becomes even more difficult, as identifying the correct representation requires manual searching by sight, since there is no way to input the radical for automated searching, as shown in Figure \ref{fig:eg-untypable}. This limitation forces us to decompose some radicals further to maintain the integrity of character representation. Unfortunately, this results in a loss of meaning in certain cases, as the radicals become fragmented beyond their functional roles within characters.

\section{Discussion on Chinese Characters}
\begin{figure}
    \centering
    \includegraphics[width=1\linewidth]{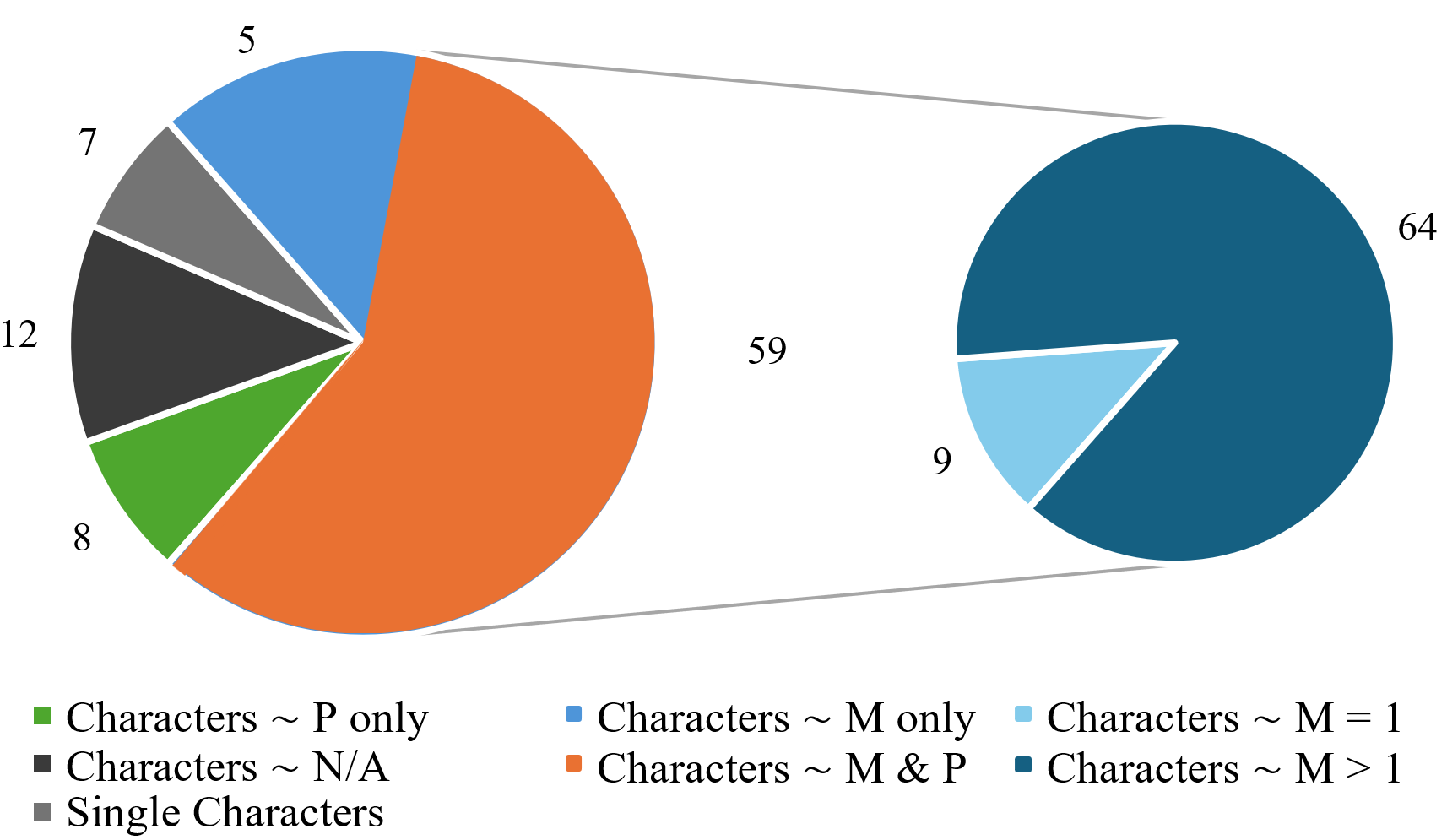}
    \caption{Distribution of Chinese characters with meaning (M) or pronunciation (P) hint from their radicals. The smaller circle on the right shows the distribution among all characters containing radicals with meaning (sum of Characters ~M only and Characters ~M \& P).}
    \label{fig:mpdist}
\end{figure}

\begin{figure}
    \centering
    \includegraphics[width=1\linewidth]{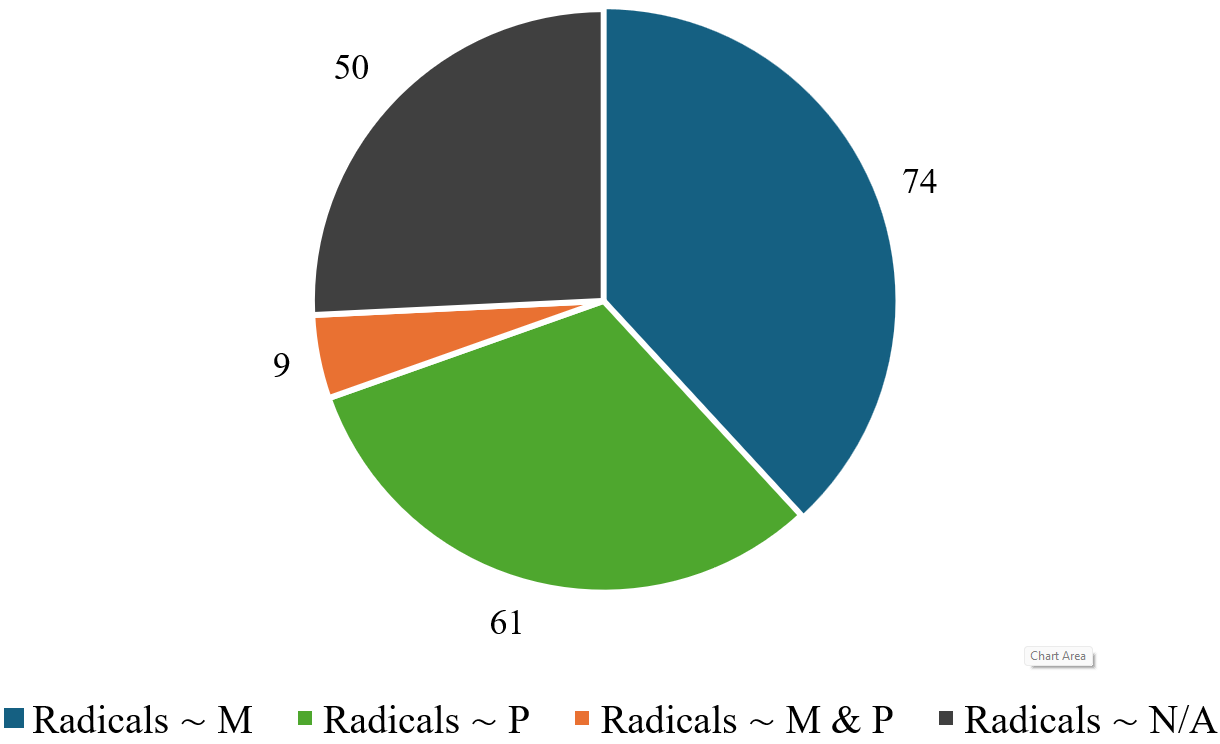}
    \caption{Sampled distribution of radicals with meaning (M) or Pronunciation (P) hint.}
    \label{fig:mpdistr}
\end{figure}

\begin{CJK*}{UTF8}{gbsn}
To investigate the importance of Chinese radicals, we selected a sample of 100 Chinese characters from our dataset and annotated them to determine whether the radicals directly contribute to the meaning or pronunciation of the character, as shown in Figures \ref{fig:mpdist}. Although the majority of characters have clues derived from the radicals, we found that most characters contain a combination of only one meaningful radical with other radicals hinting at pronunciation. For example, in the character ``花,'' we can infer that it is related to herbs from the radical ``艹,'' while ``化'' only provides a pronunciation hint, resulting in only vague idea of character's meaning. In 12 out of the 100 characters, none of the radicals were helpful. 

This is due to the evolution of the language, where historically, a single Chinese character often conveyed the meaning of a full word. However, more words are now composed of two or more characters, leading to individual characters losing their original meanings. For example, the Chinese character ``况'' is now commonly used to mean ``situation'' in words like ``情况'' or ``状况''. However, the original meaning of the character is ``cold water'' unexpectedly, which is closely related to the radical ``冫'', referring to cold water.
\end{CJK*}

\section{Detailed Radical Prompting Result}
\subsection{Quantitative Analysis on POS tagging Accuracy}
We provide a case analysis for POS tagging in Table \ref{tab:ic-count}.
\label{subsec: ic}
\begin{table}[ht]
\centering
\small
\begin{tabular}{lccc}
\toprule
\textbf{Category} & \textbf{Baseline} & \textbf{RP (Oracle)} \\ \midrule
Correct\& utilize Radical & - & 81.2 \textcolor{darkgreen}{(+81.2)}  \\
Correct without& 608.6 & 611.2 \textcolor{darkgreen}{(+2.6)}  \\
Incorrect \& utilize Radical & - & 41.8 \textcolor{red}{(+41.8)}  \\
Incorrect without & 391.4 & 265.8  \textcolor{darkgreen}{(-125.6)}  \\ \bottomrule
\end{tabular}
\caption{Quantitative analysis of GPT-3.5-Turbo's POS tagging accuracy on the number of correct and incorrect predictions with and without the examination of components using radical prompting compared to the baseline. Improvement is shown in \textcolor{darkgreen}{green}.}
\label{tab:ic-count}
\end{table}

\subsection{Window size's impact on POS tagging}
\label{subsec:windowsize}
We evaluate the impact of different window size in POS tagging with GPT-3.5-Turbo in Table \ref{tab:posablation}.
\begin{table}[h]
\centering
\setlength{\tabcolsep}{3.5pt}
\small
\begin{tabular}{l c c c}
\toprule
\multirow{3}{*}{\textbf{Window Size}} & \multicolumn{3}{c}{\textbf{Part-Of-Speech Tagging}} \\
\cmidrule(lr){2-4}
& \multicolumn{3}{c}{\textbf{GPT-3.5-Turbo with GSD}} \\
\cmidrule(lr){2-4} 
 & \textbf{B} & \textbf{RP} & \textbf{RP (Oracle)}  \\
\midrule
5 & 59.08 & 64.62 \textcolor{darkgreen}{(+5.5)} & 67.56 \textcolor{darkgreen}{(+8.5)}  \\
7 & 60.17 & 66.55 \textcolor{darkgreen}{(+6.38)} & 66.73 \textcolor{darkgreen}{(+6.56)}   \\
9 & 60.38 & 67.03 \textcolor{darkgreen}{(+6.65)} & 67.23 \textcolor{darkgreen}{(+6.85)}  \\
\bottomrule
\end{tabular}

\caption{Model performance for POS tagging with different word window sizes}
\label{tab:posablation}
\end{table}
\subsection{Radical Prompting Prompts}
\label{subsec:promptinglines}
\begin{figure*}
    \centering
    \includegraphics[width=1\linewidth]{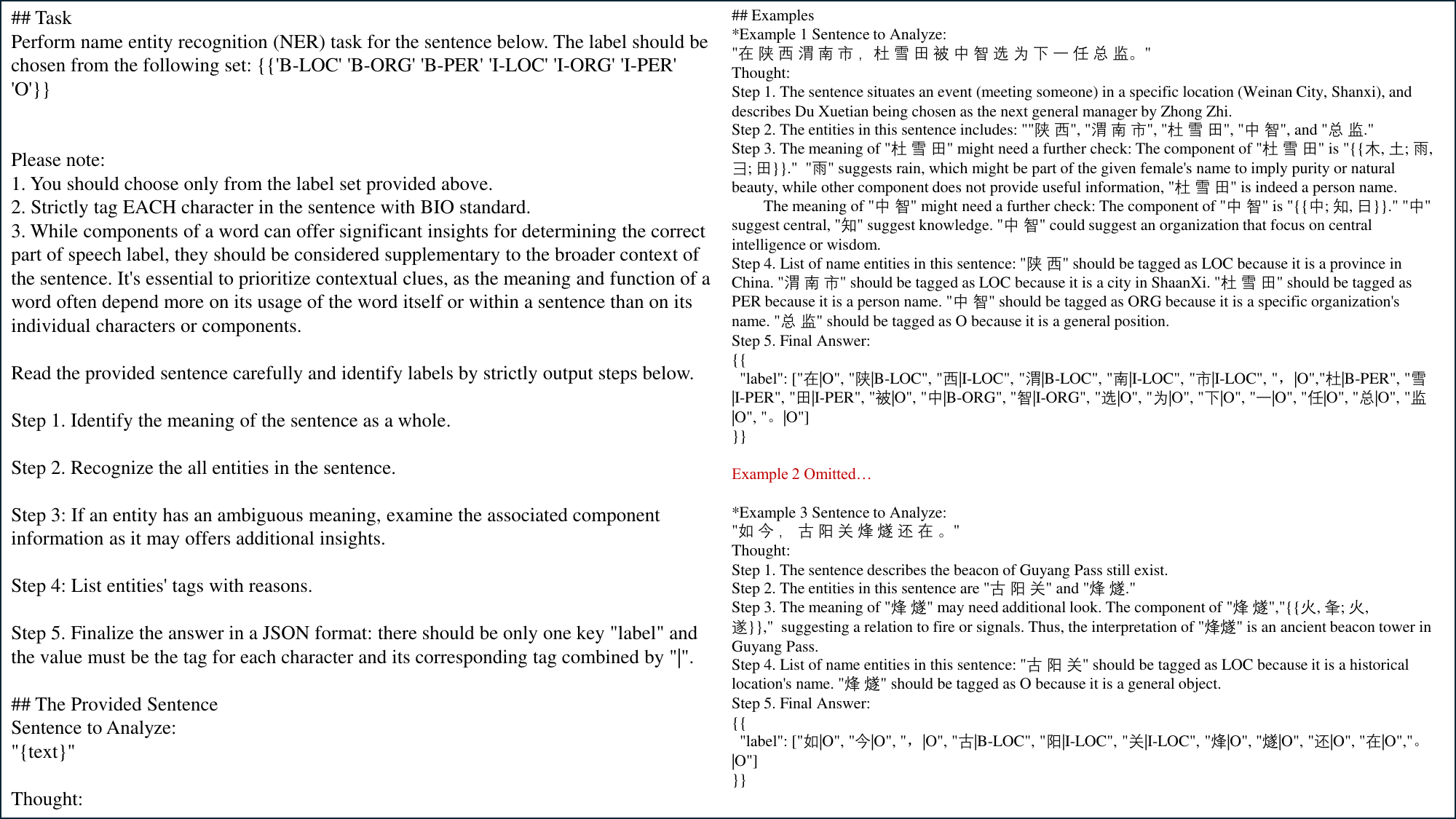}
    \caption{Radical Prompting Prompt Line of POS tagging.}
    \label{fig:promptpos}
\end{figure*}

\begin{figure*}
    \centering
    \includegraphics[width=1\linewidth]{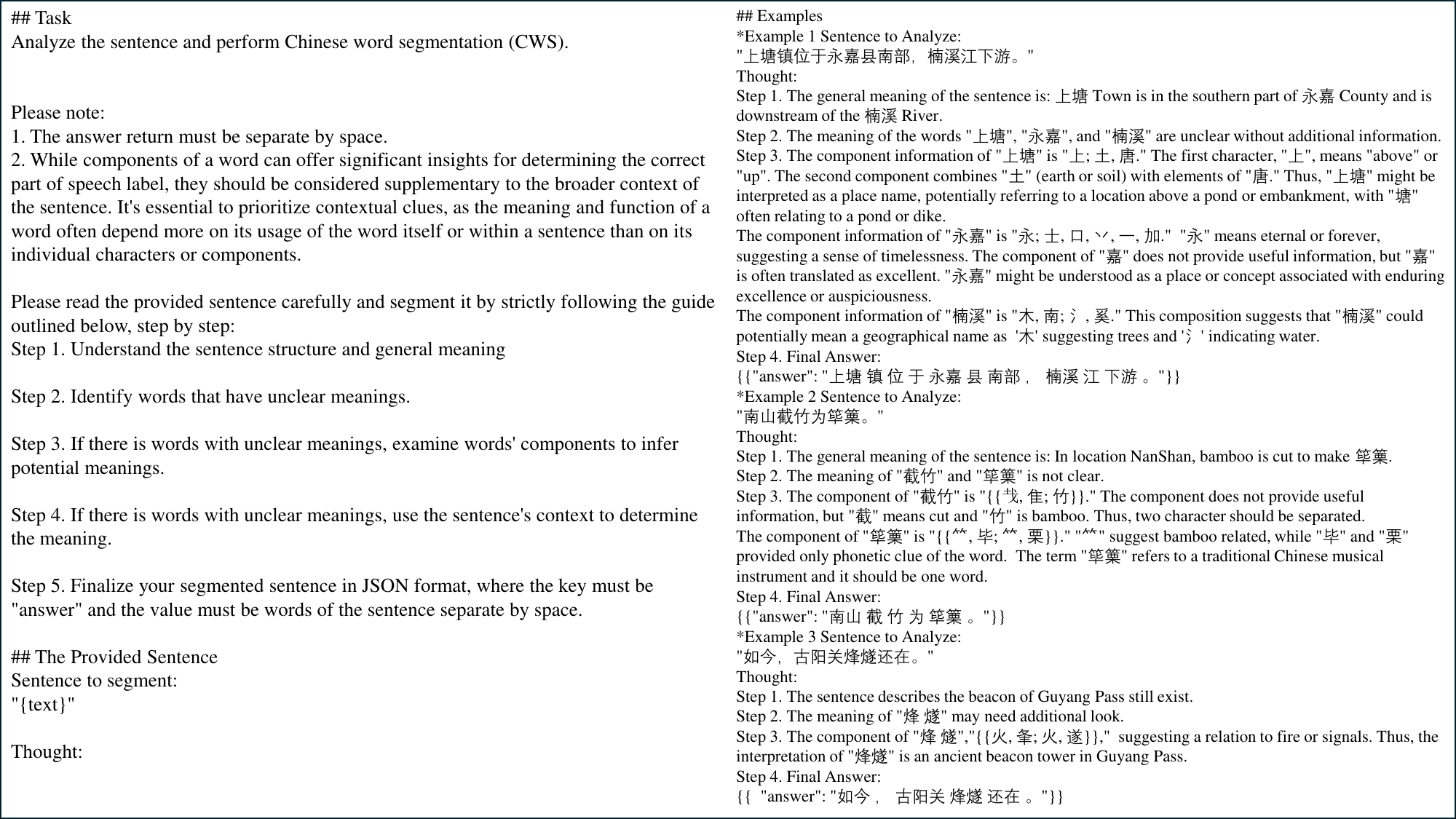}
    \caption{Radical Prompting Prompt Line of NER.}
    \label{fig:promptner}
\end{figure*}

\begin{figure*}
    \centering
    \includegraphics[width=1\linewidth]{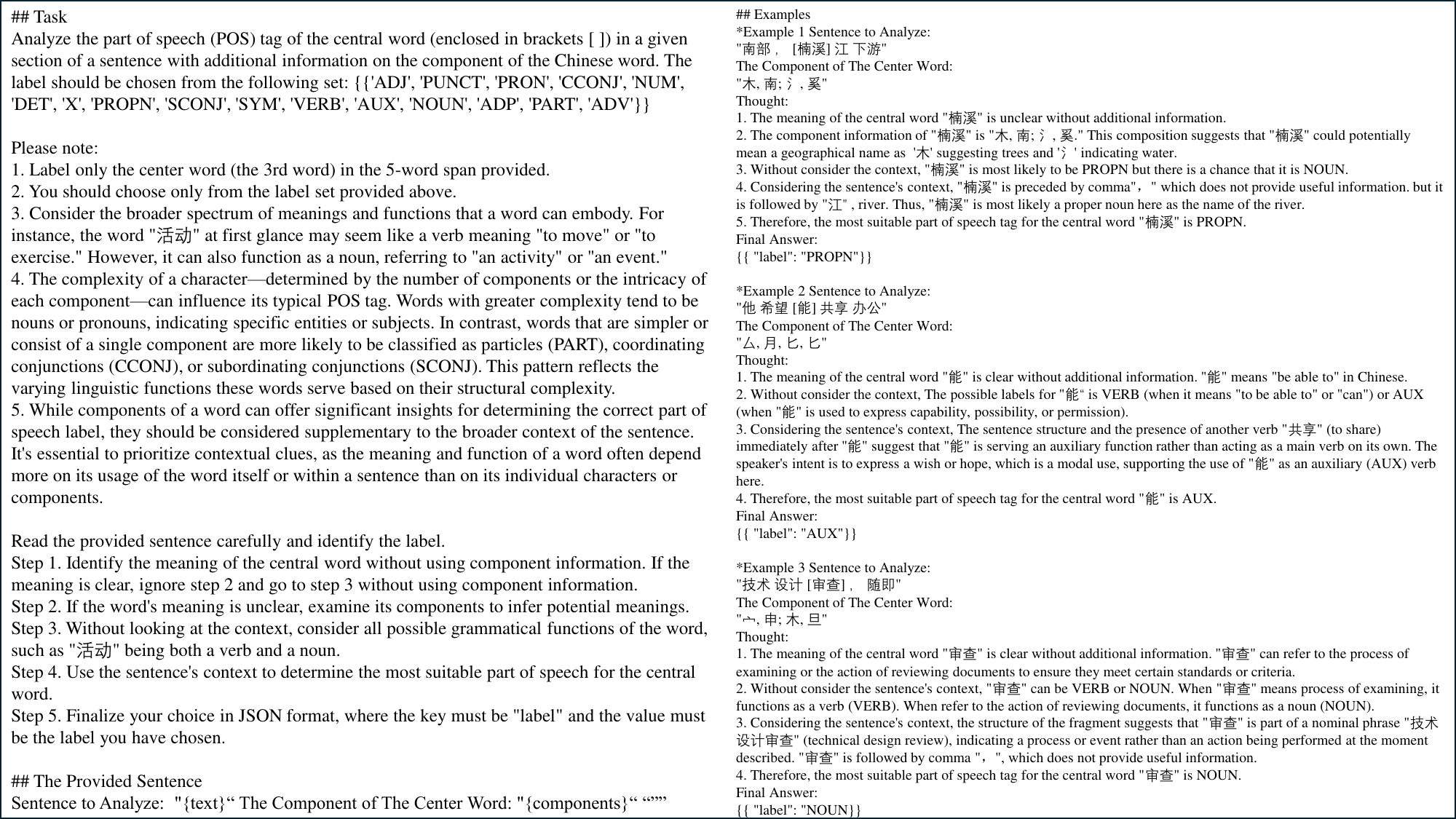}
    \caption{Radical Prompting Prompt line for CWS.}
    \label{fig:promptcws}
\end{figure*}

We provide our radical prompting lines for POS tagging, NER, and CWS tasks in Figure \ref{fig:promptpos}, \ref{fig:promptner}, and \ref{fig:promptcws}, respectively.
\subsection{Base Prompting Prompts}
\label{subsec:basepromptinglines}
\begin{figure*}
    \centering
    \includegraphics[width=1\linewidth]{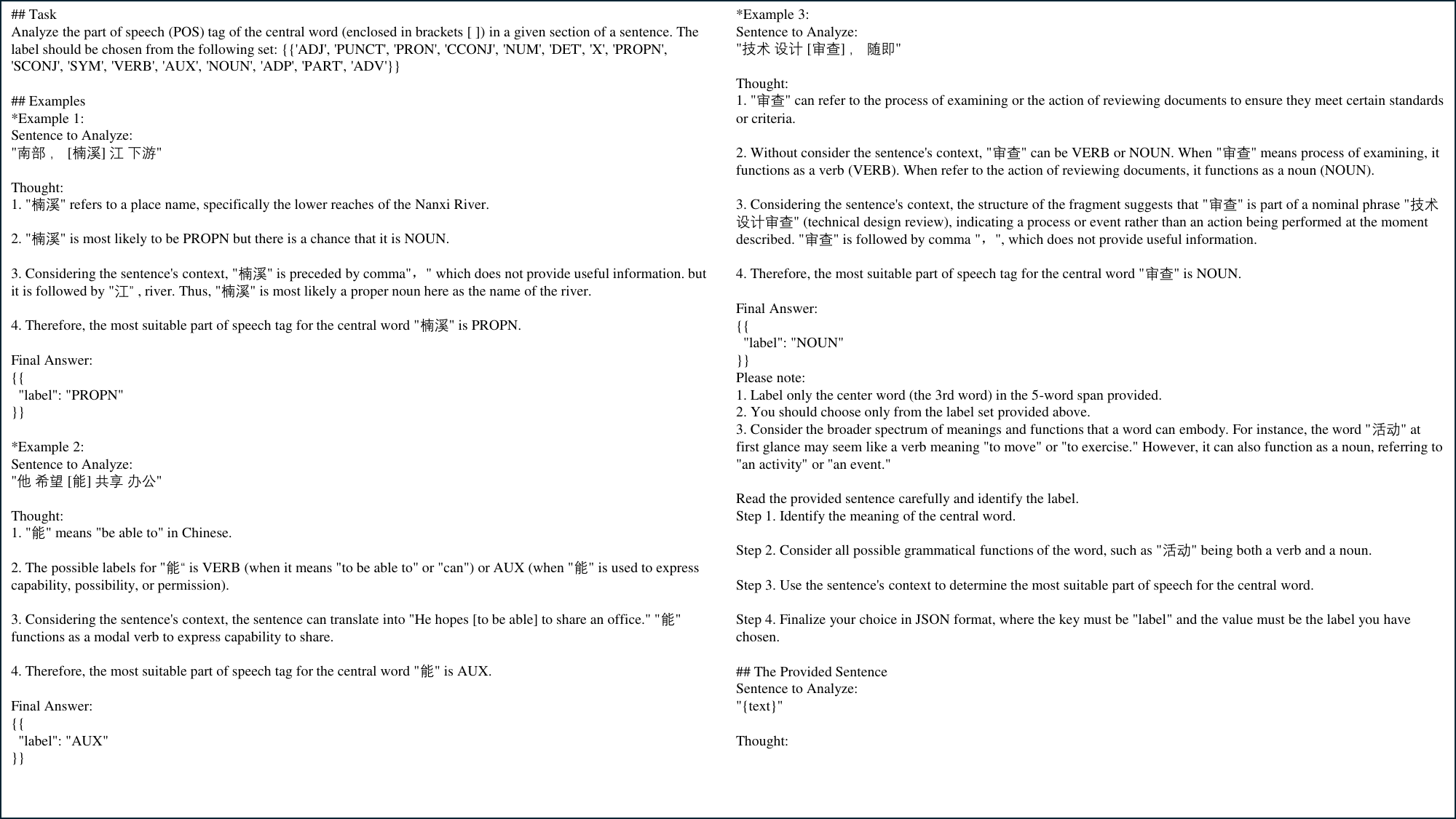}
    \caption{Base Prompt line for POS tagging.}
    \label{fig:basepromptpos}
\end{figure*}

\begin{figure*}
    \centering
    \includegraphics[width=1\linewidth]{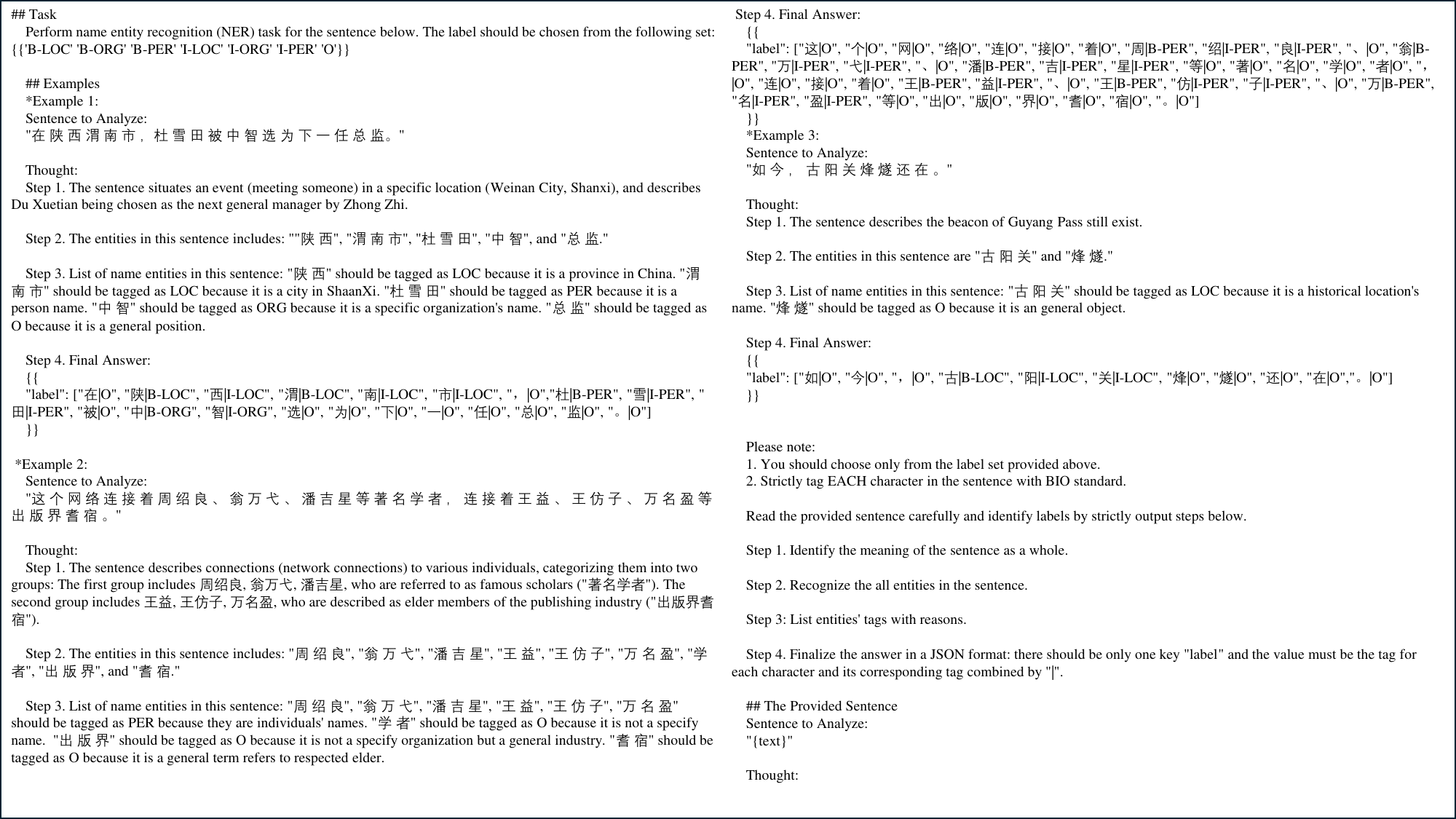}
    \caption{Base Prompt line for NER.}
    \label{fig:basepromptner}
\end{figure*}

\begin{figure*}
    \centering
    \includegraphics[width=1\linewidth]{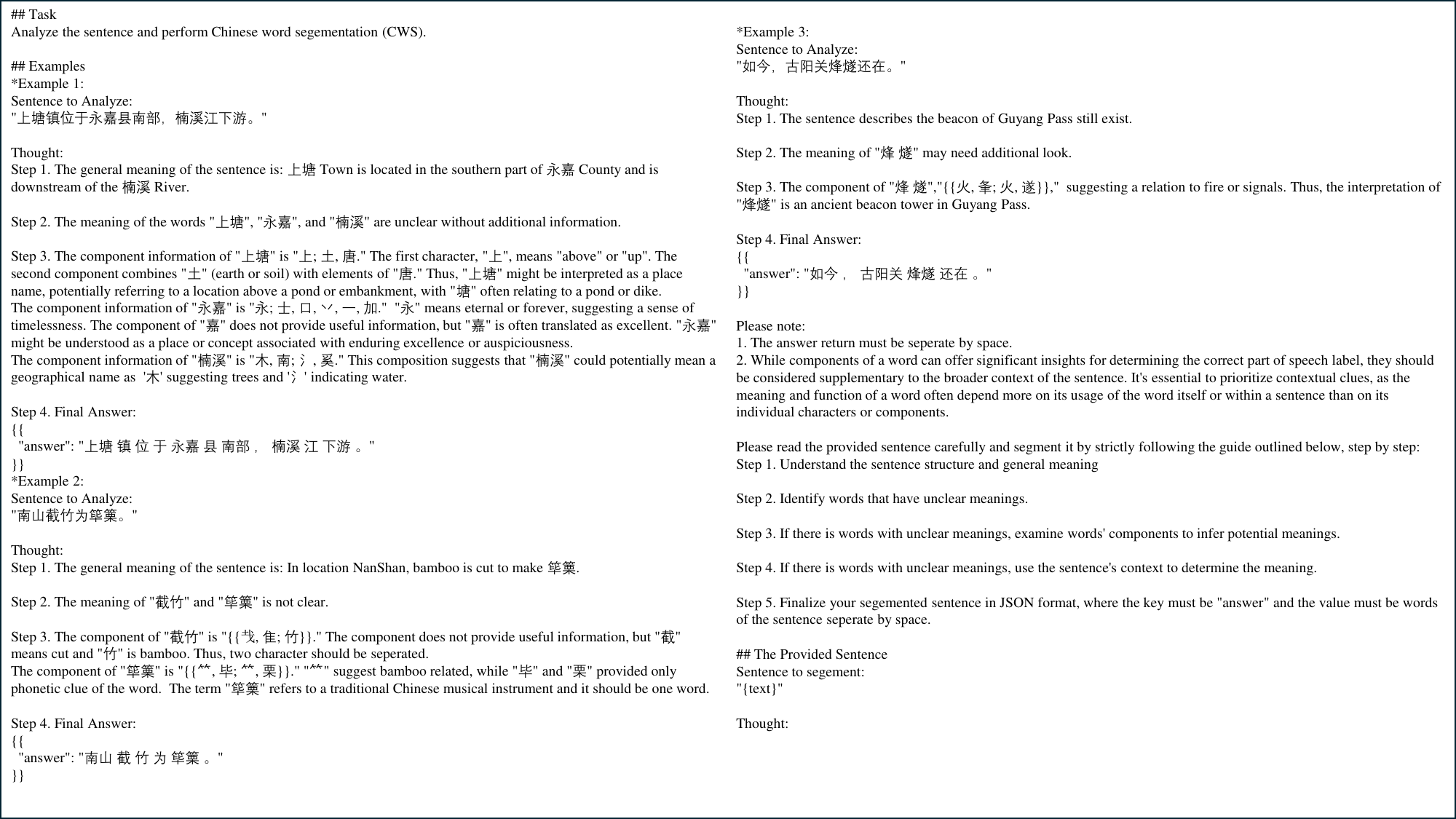}
    \caption{Base Prompt line for CWS.}
    \label{fig:basepromptcws}
\end{figure*}

We provide our base prompting lines for POS tagging, NER, and CWS tasks in Figure \ref{fig:basepromptpos}, \ref{fig:basepromptner}, and \ref{fig:basepromptcws}, respectively.

\subsection{Aya Model Behavior}
\label{ayaeg}
\begin{figure*}
    \centering
    \includegraphics[width=1\linewidth]{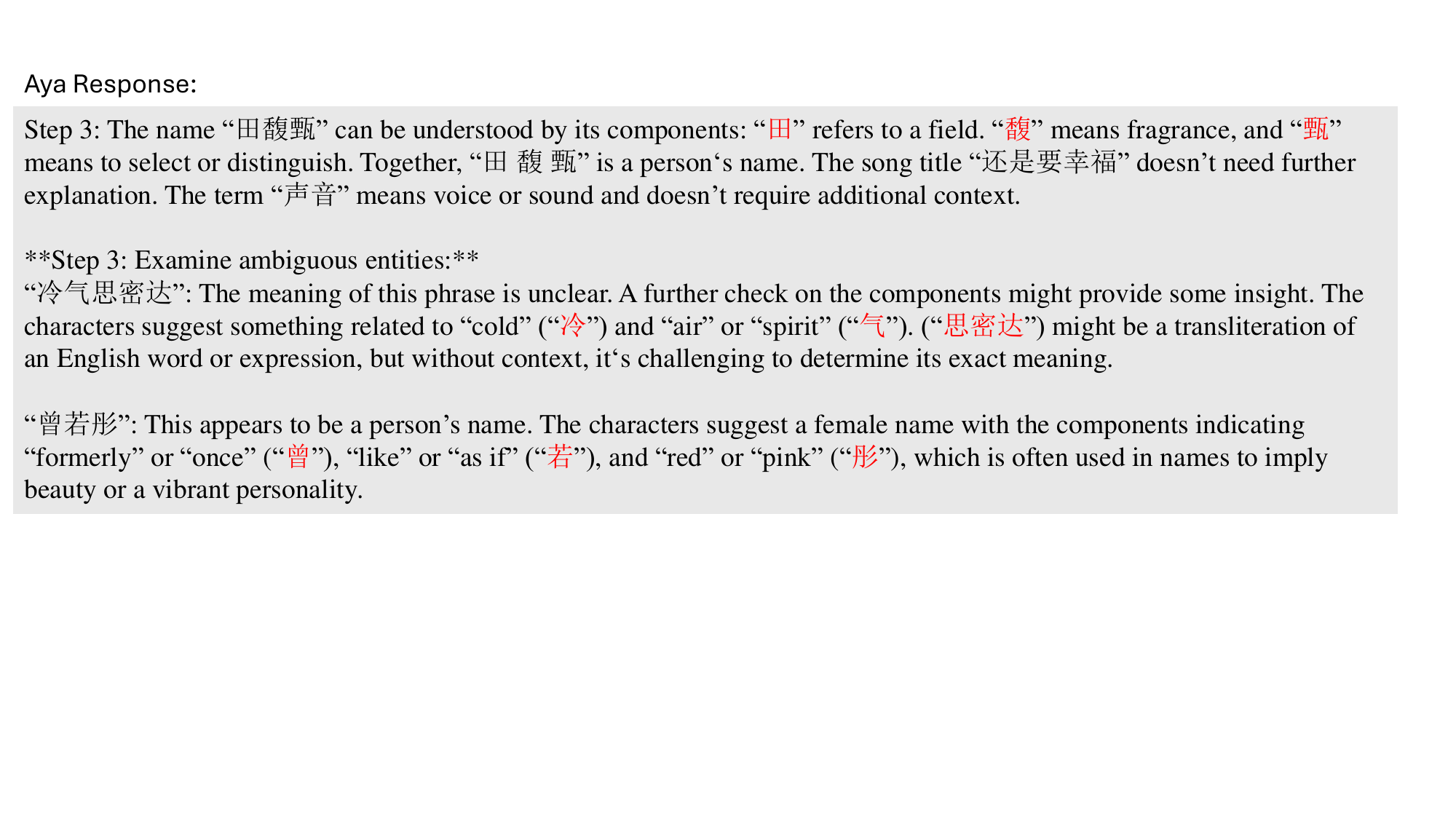}
    \caption{Example of Aya decompose incorrectly.}
    \label{fig:ayeg}
\end{figure*}

Examples of Aya decompose radicals incorrectly are shown in Figure \ref{fig:ayeg}.





\section{Responsible NLP Miscellanea}
\subsection{Intent usage}
In response to potential inquiries regarding the scope and legitimacy of our experiments, it is important to clarify that all aspects of our research strictly adhere to the intended use cases of the Large Language Models (LLMs) and the NLP task datasets employed. Furthermore, our use of these models and datasets complies fully with the usage policies of the APIs for each model involved. We note that the use of rare Chinese words triggered some safety mechanisms in models such as Gemini-1.5. However, our intent complies fully with the ethical guidelines and usage policies provided by the API providers.

\subsection{Computational Experiments Cost}

In our research, we utilized vLLMs for evaluation on Yi 6B, Mistral 7B, Baichuan 13B, and Qwen 7B with a single a40 GPU. For other models, we accessed them through their respective APIs. The cost and running time for each model varied significantly. Specifically, the time required to run a single evaluation ranged from approximately 2 to 8 hours.

\subsection{Avoid Data Leakage}
For all NLP tasks assessed in this study, evaluations were exclusively conducted on the development sets of the respective datasets to prevent data leakage.

\subsection{Personally Identifying Info}
The dataset we created for evaluating the visual information of Chinese characters does not contain any offensive content or personally identifying information. However, we acknowledge the presence of individual names in the Weibo NER dataset that we use for evaluation.

\subsection{Evaluation Tools and Methodologies}

To evaluate our Named Entity Recognition (NER) tasks, we used a Perl script: conlleval.pl.

For other tasks, we calculated F1 score using Scikit-learn. 

\subsection{AI Assistants}

We acknowledge the use of GPT-4 for grammar checking and word polishing.
\end{document}